\documentclass[]{interact}

\usepackage{epstopdf}
\usepackage[caption=false]{subfig}

\usepackage{enumerate}
\usepackage{comment}
\usepackage{CJKutf8}
\usepackage{multirow}
\usepackage{amsmath}
\usepackage{array}
\usepackage{algorithm}
\usepackage{algpseudocode}
\usepackage{lineno}
\usepackage[numbers,sort&compress]{natbib}
\bibpunct[, ]{[}{]}{,}{n}{,}{,}
\renewcommand\bibfont{\fontsize{10}{12}\selectfont}
\makeatletter
\def\NAT@def@citea{\def\@citea{\NAT@separator}}
\makeatother

\theoremstyle{plain}

\theoremstyle{definition}

\theoremstyle{remark}

\begin{document}

\articletype{FULL PAPER}

\title{Symbol emergence as interpersonal cross-situational learning: \\the emergence of lexical knowledge with combinatoriality}

\author{
\name{Yoshinobu Hagiwara\textsuperscript{a}\thanks{CONTACT Yoshinobu Hagiwara. Email: yhagiwara@em.ci.ritsumei.ac.jp},
Kazuma Furukawa\textsuperscript{a}, Takafumi Horie\textsuperscript{a}, Akira Taniguchi\textsuperscript{a}, and \\Tadahiro Taniguchi\textsuperscript{a}}
\affil{\textsuperscript{a}Ritsumeikan University \\ 1-1-1 Noji Higashi, Kusatsu, Shiga 525-8577, Japan}
}

\maketitle

\begin{abstract}
We present a computational model for a symbol emergence system that enables the emergence of lexical knowledge with combinatoriality among agents through a Metropolis-Hastings naming game and cross-situational learning. Many computational models have been proposed to investigate combinatoriality in emergent communication and symbol emergence in cognitive and developmental robotics. However, existing models do not sufficiently address category formation based on sensory-motor information and semiotic communication through the exchange of word sequences within a single integrated model.
Our proposed model facilitates the emergence of lexical knowledge with combinatoriality by performing category formation using multimodal sensory-motor information and enabling semiotic communication through the exchange of word sequences among agents in a unified model. Furthermore, the model enables an agent to predict sensory-motor information for unobserved situations by combining words associated with categories in each modality.
We conducted two experiments with two humanoid robots in a simulated environment to evaluate our proposed model. The results demonstrated that the agents can acquire lexical knowledge with combinatoriality through interpersonal cross-situational learning based on the Metropolis-Hastings naming game and cross-situational learning. Furthermore, our results indicate that the lexical knowledge developed using our proposed model exhibits generalization performance for novel situations through interpersonal cross-modal inference.
\end{abstract}


\section{Introduction}
\label{sec:introduction}
\begin{CJK}{UTF8}{min}

Understanding the phenomena that symbol systems with combinatoriality emerge through individual cognition and inter-individual communication is a crucial challenge in cognitive and developmental robotics.
Humans possess the ability to create and share words associated with cognitive categories formed from multimodal sensory-motor information~\cite{Barsalou99, SER, SEC}, and to communicate complex thoughts using word sequences as the novel combinations of words~\cite{Partee04, Zuidema18, Boer12, Compositionality}. For example, humans can form multimodal cognitive categories (e.g. green category and box category) based on the sensor information, and share word sequences representing situations such as “green box” and “red cup”. Furthermore, humans can compose a new word sequence “green cup” using a combination of the word “green” associated with the category of green color and the word “cup” associated with the category of cup object, and predict the multimodal sensor information represented by a word sequence “green cup”~\cite{Fushan19}. 
In the developmental process of human infants, it is believed that human infants learn that the word "red" represents the red color by observing the word "red" and red-colored objects in various situations, that is known as cross-situational learning (CSL)~\cite{Smith11, Fontanari09}.
As mentioned above, words associated with categories in specific modalities (e.g. color and object) allow us to represent and share various situations through the combinations of words.
We refer to such knowledge related to words associated with categories in specific modalities as lexical knowledge with combinatoriality. 

\begin{figure}
  \begin{center}			
  \includegraphics[scale=0.70]{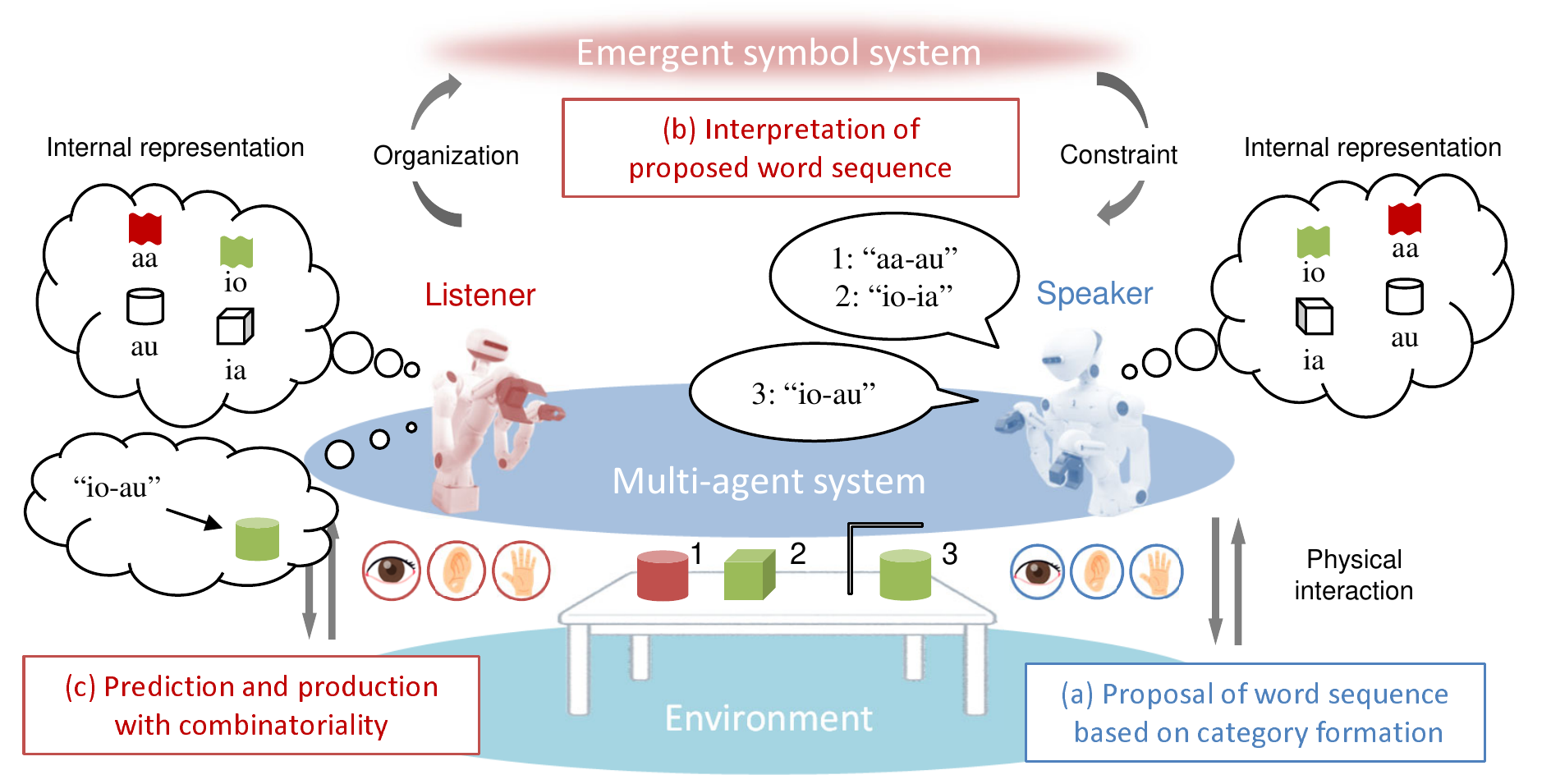}
  \caption{Overview of the symbol emergence system where lexical knowledge with combinatoriality emerges among agents; the numbers next to the objects and word sequences indicate their correspondence. While objects 1 and 2 can be observed by both agents, the listener cannot observe object 3.}
  \label{fig:em_sys}
  \end{center}
\end{figure}

This study focuses on a symbol emergence system where the lexical knowledge with combinatoriality emerges through category formation in each agent and the exchange of word sequences between agents.
This study is based on an approach of symbol emergence in robotics~\cite{Taniguchi19, Taniguchi18} which is a constructive approach toward an emergent symbol system that is socially self-organized through semiotic communications and physical interactions with autonomous cognitive developmental agents. Figure~\ref{fig:em_sys} shows an overview of the symbol emergence system where the emergence of lexical knowledge with combinatoriality among agents targeted in this study.
The main components of the symbol emergence system are as follows:
\begin{enumerate}[(a)]
\item Two agent forms categories from multimodal sensory-motor information, such as color, object, position, and action, and creates a word sequence (e.g. “aa-au”) by combining words associated with the formed categories in each modality.
\item Through repeated category formation and word sequence proposals between agents, a shared interpretation of the words associated with categories in multiple modalities is established. For instance, “io” represents the green category in the color modality, and “au” represents the box category in the object modality.
\item When a new word sequence “io-au” is proposed by one agent to represent an unobserved object (i.e. a green cup), another agent can predict the object's sensory information by combining the known words “io” and “au” associated with categories in the color and object modalities.
\end{enumerate}

Many computational studies have been conducted in the fields of cognitive and developmental robotics to understand the evolutionary and developmental dynamics of emergent communication and symbol emergence with compositionality~\cite{Kottur17, Lazaridou18, Mordatch18, Cogswell19, Nefdt20, Chaabouni20, Eccles19, Simoes19, Cowen20, Lazaridou20}.
Nefdt analyzed the compositionality of language by assuming full knowledge of primitive words and their combination rules~\cite{Nefdt20}. However, lack of analysis of compositionality with the emergence of language.
Kottur demonstrated that emergent language could help to correctly refer to unseen composite inputs in multi-agent cooperative tasks~\cite{Kottur17}, and the generalization performance of language was analyzed. However, no model explaining how this performance arises was provided.
Many studies based on deep reinforcement learning have been conducted to model emergent communication~\cite{Eccles19, Simoes19, Cowen20, Lazaridou20}. 
Chaabouni et al. demonstrated that neural agents could generalize to unseen combinations of inputs in a simple communication game~\cite{Chaabouni20}.
However, these studies assumed a conceptualized simulation environment and did not include a model for how agents acquire concepts and categories from their sensory-motor information.

Computational models, inspired by human language acquisition, have been proposed for robotic learning of word meanings associated with concepts and categories from sensory-motor information~\cite{Nakamura09, Nakamura14, Muhammad13, Miyazawa19, Ishibushi15, Taniguchi17, Hagiwara18, Akira20, CSL}. 
Nakamura et al. developed a probabilistic generative model (PGM) that empowers a robot to acquire concepts and associated language based on multimodal sensory information (e.g., vision, haptic, sound, and linguistic information)~\cite{Nakamura14}. 
Taniguchi et al. extended this model to a PGM for cross-situational learning (CSL-PGM). The robot learns to associate categories acquired from each modality with word sequences provided by a human tutor~\cite{CSL}. The CSL-PGM enables an agent to understand the correspondence between words in sentences and categories in multiple modalities from given word sequences and categories formed from sensory-motor information. However, these models rely on learning with a human tutor and do not account for symbol emergence between agents.

As studies on computational models for symbol emergence with concept acquisition, many models have been proposed in developmental and evolutionary systems, including robotics~\cite{Steels95, Steels06, Steels14, Spranger11, Spranger15, Vogt02, Vogt05, Beule06, Bleys15, Matuszek18}. 
Steels et al. modeled the process by which agents acquire concepts of simple-shaped objects from real-world data and the emergence of symbol systems through the exchange of words between agents as a language game~\cite{Steels95, Steels06, Steels14}.
The language game is a popular approach for modeling symbol emergence~\cite{Steels15}, with naming games being a representative example among language games. These naming games allow multiple agents to communicate with each other, form object categories, and share names (i.e. words).
However, their models do not address the formation of categories and the emergence of symbols as a single integrated model.

Hagiwara et al. proposed a naming game based on a PGM with the Metropolis-Hastings method~\cite{Hastings70}, which is a type of Markov Chain Monte Carlo algorithm~\cite{Inter-DM}. This naming game has been referred to as the Metropolis-Hastings naming game (M-H naming game)~\cite{Taniguchi22}.
The model used in this game is the interpersonal multimodal Dirichlet mixture (Inter-MDM), which combines two PGMs for multimodal categorization by two agents into a single PGM and enables agents to perform multimodal categorization and naming game~\cite{Inter-MDM}.
The Metropolis-Hastings method ensures that word sharing between agents is possible in a probabilistic manner and the process of symbol emergence is viewed as an inference process of the shared variable (i.e., the index of a word) from a Bayesian perspective.
However, the Inter-MDM model was limited to communication using a single word and could not handle communication based on word sequences comprising multiple words.
This study extends the model of an M-H naming game~\cite{Inter-MDM} with the CSL-PGM~\cite{CSL} to address the combinatoriality in lexical knowledge.

We present a novel computational model that extends the M-H naming game model~\cite{Inter-MDM-H2H} with CSL-PGM~\cite{CSL} to represent the symbol emergence system illustrated in Figure~\ref{fig:em_sys}. Our proposed model employs a CSL model for each agent, allowing the M-H naming game to be based on a word sequence instead of a single word. Through the repeated play of the game, the agents learn the cross-situational relationships between words in word sequences and categories across multiple modalities. By combining these learned words associated with categories in each modality, the proposed model enables the agents to generate word sequences that describe novel situations and to predict sensory-motor information from new word sequences.

To evaluate our proposed model, we conducted two experiments in a simulation environment using two humanoid robots.
In the first experiment, two agents played the M-H naming game in various situations and shared lexical knowledge through interpersonal cross-situational learning. The experiment clarifies whether lexical knowledge with combinatoriality emerges among agents in our proposed model by analyzing and evaluating the formed categories and shared words.
In the second experiment, one agent predicted sensory-motor information from new word sequences proposed by another agent in novel situations through interpersonal cross-modal inference. The experiment clarifies whether the proposed model enables agents to predict multimodal sensory-motor information of novel situations from the new combination of words associated with each modality.

The main contributions of this study are as follows:
\begin{itemize}
\item We proposed a computational model for a symbol emergence system that allows the emergence of lexical knowledge with combinatoriality among two agents performing an M-H naming game with CSL. Our experiments empirically confirmed that the M-H naming game enables the emergence of not only categorical signs but also lexical knowledge with combinatoriality dependent on modalities. 
Furthermore, we have demonstrated that lexical knowledge facilitated by our proposed model has generalization performance to novel situations through interpersonal cross-modal inference.
\item We demonstrated that the mutual exclusivity bias (constraint), which assigns novel words to unknown situational elements, plays a crucial role in the M-H naming game with CSL for the emergence of lexical knowledge with combinatoriality.
\end{itemize}

\end{CJK}

\section{Proposed model: Interpersonal CSL-PGM (Inter-CSL-PGM)}
\label{sec:proposed}
\begin{CJK}{UTF8}{min}
We proposed a Bayesian generative model for a symbol emergence system, in which lexical knowledge with combinatoriality emerges among two agents through a M-H naming game and CSL.
Our proposed model was developed by extending the model of M-H naming game~\cite{Inter-MDM-H2H} with CSL-PGM~\cite{CSL}. The basic theory and models of the M-H naming game is described in Appendix~\ref{apnd:naming game}.
This section describes the generative process, inference algorithm, and prediction process in our proposed model.

\subsection{Generative process}

\begin{figure}
  \begin{center}			
  \includegraphics[scale=0.4]{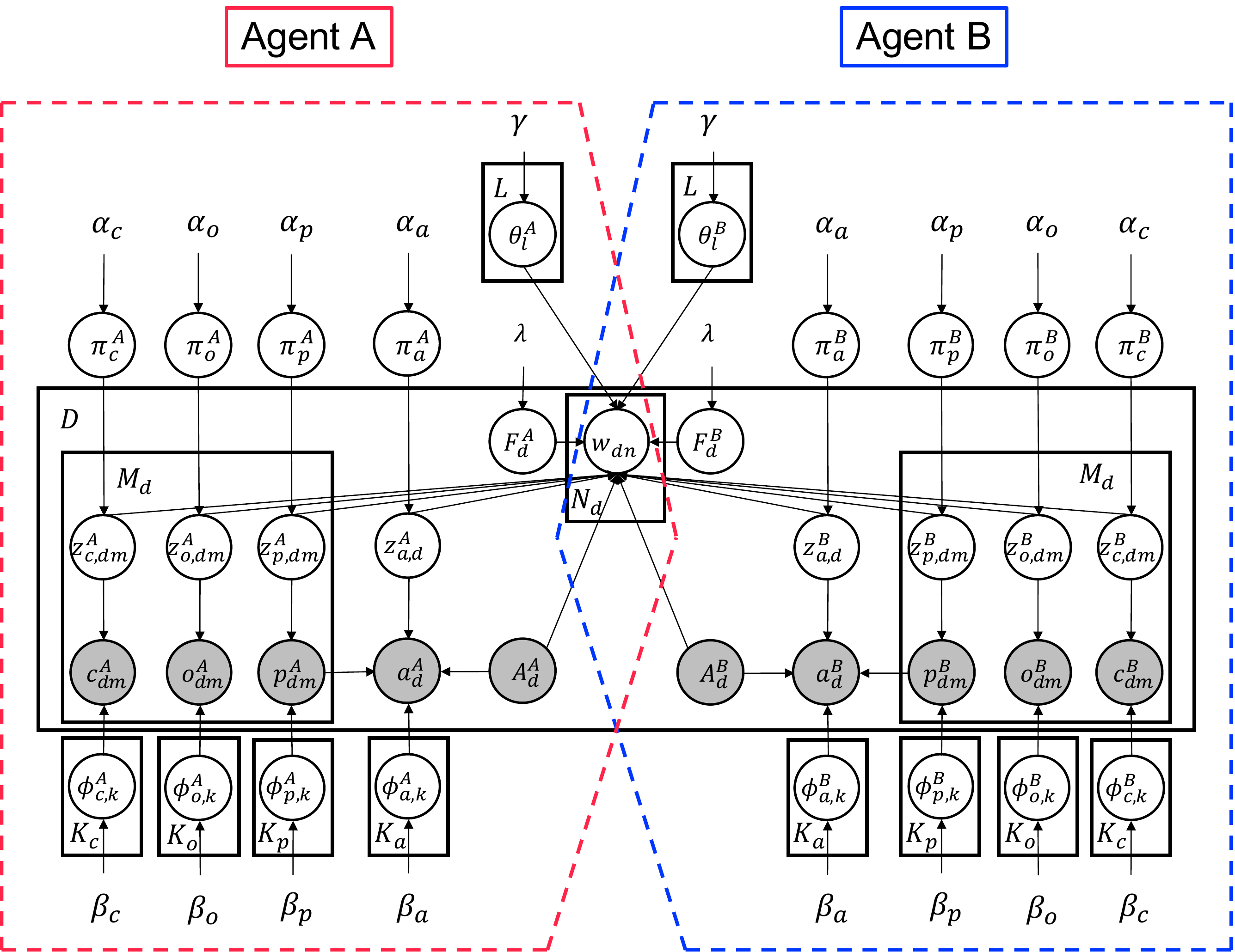}
  \caption{Graphical model of Inter-CSL-PGM; red and blue dashed lines show agents A and B, respectively.}
  \label{fig:multi_csl}
  \end{center}
\end{figure}

\begin{table}[tb!]
  \begin{center}
  \caption{Definition of variables in Inter-CSL-PGM. $*$ denotes one of the modalities, i.e., a: action, p: position, o: object, and c: color. As the structure of agents A and B are identical, the subscripts to denote the agents were omitted.}
    \footnotesize
    \renewcommand{\arraystretch}{1.2}
    \begin{tabular}{c l} 
      Variables & Definition \\ \hline
      $D$ & Number of trials \\
      $K_*$ & Number of categories in a modality ($*$) \\
      $L$ & Sum of the number of categories ($L=K_a+K_p+K_o+K_c$) \\
      $M_d$ & Number of objects on the table in $d$-th trial \\
      $N_d$ & Number of words in the word sequence in $d$-th trial \\
      $A_d$ & Index ($m$) of the object of attention selected from multiple objects on the table \\
      $F_d$ & Parameter to determine the order of modalities in a word sequence \\
      $a_d$ & Observation of action modality \\
      $p_{dm},o_{dm},c_{dm}$ & Observations of position, object, and color modalities for $m$-th object \\
      $z_{a,d}$ & Index of the category of action modality \\
      $z_{p,dm}$, $z_{o,dm}$, $z_{c,dm}$ & Indices of the categories of position, object, and color modalities for $m$-th object \\
      $\pi_*$ & Parameter for generating the index of a category ($z_{*}$) \\
      $\phi_{*,k}$ & Parameter for generating an observation ($*_d$) in $k$-th category\\
      $w_{dn}$ & $n$-th word in a word sequence \\ 
      $\theta_l$ & Parameter for generating a word ($w_{dn}$) from the index ($l$) of a category with $F_d$ \\
      $\alpha_*,\beta_*,\gamma,\lambda$ & Hyperparameters for $\theta,\phi,w_d$ \\  \hline
    \end{tabular}
  \label{tab:PGM}
  \end{center}
\end{table}

Figure~\ref{fig:multi_csl} shows the graphical model of our proposed PGM. The PGM consists of two CSL-PGMs~\cite{CSL}, represented by Agents A and B, which are connected by a word sequence ($w_{d}$) as a connection variable inferred by the M-H naming game. We refer to this PGM as Inter-CSL-PGM (Interpersonal CSL-PGM).
Table~\ref{tab:PGM} shows the definition of variables in Inter-CSL-PGM.

As the structure of agents A and B are identical, the generative process that is common among the agents only is explained for Agent A.
Equations (\ref{eq:Fd})-(\ref{eq:ad}) show the generative process.
{\small
\begin{equation}
    \label{eq:Fd}
    F_d^A \sim {\rm Unif}(\lambda)
\end{equation}
\begin{equation}
    \label{eq:thetal}
    \theta_{l}^A \sim {\rm Dir}(\gamma)
\end{equation}
\begin{equation}
    \label{eq:pi}
    \pi^A_{*} \sim {\rm GEM}(\alpha_{*})
\end{equation}
\begin{equation}
    \label{eq:zdm}
    z_{*,dm}^A \sim {\rm Cat}(\pi^A_{*})
\end{equation}
\begin{equation}
    \label{eq:phik}
    \phi^A_{*,k} \sim {\rm GIW}(\beta_{*})
\end{equation}
\begin{equation}
    \label{eq:odm}
    o_{dm}^A \sim {\rm Gauss}(\phi_{o,z_{o,dm}^A}^{A})
\end{equation}
\begin{equation}
    \label{eq:cdm}
    c_{dm}^A \sim {\rm Gauss}(\phi_{c,z_{c,dm}^A}^{A})
\end{equation}
\begin{equation}
    \label{eq:pdm}
    p_{dm}^A \sim {\rm Gauss}(\phi_{p,z_{p,dm}^A}^{A})
\end{equation}
\begin{equation}
    \label{eq:ad}
    a_{d}^A \sim {\rm Gauss}(\phi_{a,z_{a,d}^A}^{A})
\end{equation}
}

In the generative process, $\rm{Unif}(\cdot)$ denotes the uniform distribution, $\rm{Dir}(\cdot)$ denotes the Dirichlet distribution, $\rm{Cat}(\cdot)$ denotes the categorical distribution, $\rm{GIW}(\cdot)$ denotes the Gaussian Inverse Wishart distribution, $\rm{Gauss}(\cdot)$ denotes the multivariate Gaussian distribution, $\rm{GEM}(\cdot)$ denotes the stick-breaking process~\cite{sbp94}. 
In Equation (\ref{eq:Fd}), $F_d$ is generated from a uniform distribution. Similar to CSL-PGM~\cite{CSL}, the tree structure or order of words in a word sequence is not learned in Inter-CSL-PGM. Inter-CSL-PGM estimates cross-situations between words in a word sequence and categories in multiple modalities. This enables an agent to generate a word sequence representing a novel situation by combining words associated with categories in multiple modalities and to predict the sensory-motor information from a novel word sequence by combining categories estimated from each word in a word sequence. Details of these functions are described in Sections \ref{sec:inference} and \ref{sec:interpersonal coross-modal inference}.
In Equations (\ref{eq:odm}) to (\ref{eq:ad}), the observations ($o_{dm}, c_{dm}, p_{dm}, a_d$) in each modality is generated from a multivariate Gaussian distribution with a parameter ($\phi$).
The $\phi$ is defined as a parameter that generates observations based on category ($z$).

Equation (\ref{eq:wdn*}) is the generative process of a word ($w_{dn}$) as a connection variable between the models of two agents. 

{\small
\begin{eqnarray}
\label{eq:wdn*}
w_{dn} &\sim& P\left(w_{dn}|F_{dn}^{A},A_d^A,z_{**,dm}^{A},{\rm \bf{\Theta}}^{A},F_{dn}^{B},A_d^B,z_{**,dm}^{B},{\rm \bf{\Theta}}^{B}\right) \nonumber \\
&\approx& \hat{P}\left(w_{dn}|F_{dn}^{A},A_d^A,z_{**,dm}^{A},{\rm \bf{\Theta}}^{A},F_{dn}^{B},A_d^B,z_{**,dm}^{B},{\rm \bf{\Theta}}^{B}\right) \nonumber \\
&\propto& P\left(w_{dn}|F_{dn}^{A},A_d^A,z_{**,dm}^{A},{\rm \bf{\Theta}}^{A}\right)P\left(w_{dn}|F_{dn}^{B},A_d^B,z_{**,dm}^{B},{\rm \bf{\Theta}}^{B}\right),
\end{eqnarray}
}
where $\bf{\Theta}$ is a set of $\theta$ values, and $\hat{P}$ is an approximately decomposed distribution. This approximation is based on the product of expert~\cite{poe}, which is a method that approximates the probability distribution of observed data by a product of individual probability distributions. The details of the approximate decomposition are described in \cite{Taniguchi20}. 
In Equation (\ref{eq:wdn*}), $**$ in $z_{**,dm}$ has subscripts ($a, o, p, c$) for all modalities. 
Each modality has a latent variable representing the index of the category ($z_{*,d}$), and a word ($w_{dn}$) is generated based on the order of modalities in $F_d$.
For instance, when the number of words in a word sequence is set to $N_d=4$ and the order of modalities in $F_d$ is $\{a,c,p,o\}$, the word sequence $w_{d}$ is generated from $\theta$ corresponding to the index of a category ($z_{*,d}$) referenced in the order of action, color, position, and object modalities.
This assigns attributes of multiple modalities to the word sequence.
Attributes for multiple modalities in a word sequence enable the agent to generate a novel word sequence and predict a novel situation through the combinations of words associated with categories in each modality.

\subsubsection*{\textbf{Mutual exclusivity constraint (MEC):}}
$P(w_{dn}|F_{dn}^{A},A_d^A,z_{**,dm}^{A},{\rm \bf{\Theta}}^{A})$ and $P(w_{dn}|F_{dn}^{B},A_d^B,z_{**,dm}^{B},{\rm \bf{\Theta}}^{B})$ in Equation (\ref{eq:wdn*}) are calculated by the following equations.
{\small
\begin{eqnarray}
 \label{eq:rescaling}
 P\left(w_{dn}|F_{dn}^{A},A_d^A,z_{**,dm}^{A},{\rm \bf{\Theta}}^{A}\right) &\propto& \frac {{\rm Cat}\Bigl(\theta_{l=(F_{dn}^A,z_{F_{dn}^A,dA_{d}^A}^{A})}^A\Bigr)}{\sum_{F^{\prime},z^{\prime}}{\rm Cat}\Bigl(\theta_{l=(F_{dn}^{\prime A},z_{F_{dn}^{\prime A},dA_{d}^A}^{\prime A})}^A\Bigl)}, \nonumber \\
 P\left(w_{dn}|F_{dn}^{B},A_d^B,z_{**,dm}^{B},{\rm \bf{\Theta}}^{B}\right) &\propto& \frac {{\rm Cat}\Bigl(\theta_{l=(F_{dn}^B,z_{F_{dn}^B,dA_{d}^B}^{B})}^B\Bigr)}{\sum_{F^{\prime},z^{\prime}}{\rm Cat}\Bigl(\theta_{l=(F_{dn}^{\prime B},z_{F_{dn}^{\prime B},dA_{d}^B}^{\prime B})}^B\Bigl)}.
\end{eqnarray}
}
The denominator in Equations~(\ref{eq:rescaling}) is a re-scaling term that enables each agent to generate a variety of words. This re-scaling term is inspired by the mutual exclusivity bias~\cite{Markman89} in which infants assign novel words to unknown objects in lexical acquisition. We named this team mutual exclusivity constraint (MEC).
The numerator of Equation~(\ref{eq:rescaling}) represents the probability of generating a word ($w_{dn}$) from the categorical distribution with a parameter ($\theta$) in the category ($l$). $\theta$ is a parameter representing the probability of observing a word ($w_{dn}$) in the category ($l$). In the learning of $\theta$, a specific word that is coincidentally observed in a certain category may become more likely to be generated, raising the generation probability of a specific word across all categories in CSL.
On the other hand, the denominator is the sum of the parameters ($\theta$) for a word ($w_{dn}$) in all categories. This value represents $P(w)$ which is the probability of observing the word ($w_{dn}$). By dividing the numerator with this value, the generation probability of frequently observed words decreases, while the generation probability of less observed words increases. 
Through this re-scaling computation, we modeled the mutual exclusivity bias. It is anticipated that with this model, a new word will be allocated to a category where word learning is not progressing.

\subsection{Inference as interpersonal cross-situational learning}
\label{sec:inference}

\begin{algorithm}
 \caption{Metropolis-Hastings Naming Game}
 \label{alg:MHNG}
 \footnotesize
 \begin{algorithmic}[1]
 {
 \State {Initialize all parameters}
 \For {$t=1$ to $T$} 
  \State  //  Agent A talks to Agent B.
  \For {$d=1$ to $D$}
   \For {$n=1$ to $N_d$}
    \State { $w_{dn}^B \leftarrow$ MH-communication $(F_{dn}^{A}, A_{d}^{A}, z_{**,dm}^{A}, {\bf{\Theta}}^{A}, F_{dn}^{B}, A_d^B, z_{**,dm}^{B}, {\bf{\Theta}}^{B}, w_{dn}^B)$}
   \EndFor
  \EndFor
  \State // Learning by Agent B 
  \For {each, $* \in \{a,p,o,c\}$}
   \State $\pi_{*}^{B} \sim P(\pi_{*}^B|z_{*}^B, \alpha)$
   \State $\phi_{*}^{B} \sim P(\phi_{*}^B|*^B, \beta_*)$
   \State $\theta^{B} \sim P(\theta^{B}|w^B,\gamma)$
  \EndFor
  \State // Perception by Agent B 
  \For {$d=1$ to $D$}
   \For {each, $* \in \{a,p,o,c\}$}
    \State  $z_{*,d}^{B} \sim  P(z_{*,d}^B|*_d^B, w_d^B, \pi_{*}^B, \phi_{*}^B, \theta^B, F_d^B, A_d^B)$
   \EndFor
  \EndFor
  \State  // Agent B talks to Agent A. 
  \For {$d=1$ to $D$}
   \For {$n=1$ to $N_d$}
    \State { $w_d^A \leftarrow$ MH-communication $(F_{dn}^{B}, A_{d}^{B}, z_{**,dm}^{B}, {\bf{\Theta}}^{B}, F_{dn}^{A}, A_d^A, z_{**,dm}^{A}, {\bf{\Theta}}^{A}, w_{dn}^A)$}
   \EndFor
  \EndFor
  \State // Learning by Agent A
  \For {each, $* \in \{a,p,o,c\}$}
   \State $\pi_{*}^{A} \sim P(\pi_{*}^A|z_{*}^A, \alpha)$
   \State $\phi_{*}^{A} \sim P(\phi_{*}^A|*^A, \beta_*)$
   \State $\theta^{A} \sim P(\theta^{A}|w^A,\gamma)$
  \EndFor
  \State // Perception by Agent A
  \For {$d=1$ to $D$}
   \For {each, $* \in \{a,p,o,c\}$}
    \State  $z_{*,d}^{A} \sim  P(z_{*,d}^A|*_d^A, w_d^A, \pi_{*}^A, \phi_{*}^A, \theta^A, F_d^A, A_d^A)$
   \EndFor
  \EndFor
 \EndFor
}
\end{algorithmic} 
\end{algorithm}

To model the emergence of lexical knowledge based on the exchange of word sequences between two agents performing category formation from sensory-motor information, M-H naming game~\cite{Taniguchi22} was used to infer the parameters of Inter-CSL-PGM.
Algorithm~\ref{alg:MHNG} shows the M-H naming game in Inter-CSL-PGM. 
In Lines 4-8, the function of MH-communication(.) is used to sample and probabilistic acceptance of word sequences with agent A as the speaker and agent B as the listener.
In Lines 10-12, the parameters in Agent B are updated.
In Line 15, the index of categories in Agent B is updated.
In Lines 18-22, the function of MH-communication(.) is used to sample and probabilistic acceptance of word sequences with agent B as the speaker and agent A as the listener.
In Lines 23-30, the parameters and the index of categories in Agent A are updated.
This process can be interpreted as communication through the exchange of words between agents.
The variables denoted with $**$ are shorthand for the four variables with $a,p,o,c$. $*$ denotes one of $a,p,o,c$.

\begin{algorithm}
 \caption{Metropolis-Hastings Communication}
 \label{alg:MH-C}
 \footnotesize
 \begin{algorithmic}[1]
 {
 \Function{MH-communication}{$F_{dn}^{Sp}, A_{d}^{Sp}, z_{**,dn}^{Sp}, {\bf{\Theta}}^{Sp}, F_{dn}^{Li}, A_d^{Li}, z_{**}^{Li}, {\bf{\Theta}}^{Li}, w_{dn}^{Li}$}
 \State $w_{dn}^{Sp} \sim P(w_{dn}^{Sp}|F_{dn}^{Sp},A_{d}^{Sp},z_{**,dn}^{Sp},{\rm \bf{\Theta}}^{Sp})$
 \State $r = {\rm min}\left(1,
    \frac{
    P(w_{dn}^{Sp}|F_{dn}^{Li},A_{d}^{Li},z_{**,dn}^{Li},{\rm \bf{\Theta}}^{Li})}
    {
    P(w_{dn}^{Li}|F_{dn}^{Li},A_{d}^{Li},z_{**,dn}^{Li},{\rm \bf{\Theta}}^{Li})         
    }
    \right)$
 \State $u \sim {\rm Unif}(0,1)$
   \If {$u\leq r$}
  \State {\bf return} $w_d^{Sp}$
  \Else
  \State {\bf return} $w_d^{Li}$
  \EndIf
 \EndFunction
}
\end{algorithmic} 
\end{algorithm}

Algorithm~\ref{alg:MH-C} denotes the Metropolis-Hastings communication (MH-communication) part of Algorithm~\ref{alg:MHNG}. $Sp$ denotes the speaker, $Li$ denotes the listener, and Unif(.) denotes the uniform distribution.
The inference process of $w_{dn}$ based on the M-H naming game is interpreted as a process in which each agent probabilistically accepts a word proposed by another agent based on its own knowledge, and updates its own knowledge if the word is accepted.
Probabilistic acceptance refers to the decision to accept or reject a sampled word ($w_{dn}^{Sp}$) based on the acceptance ratio ($r$) which is calculated in Line 3 of Algorithm~\ref{alg:MH-C}. The derivation of the acceptance ratio ($r$) is described in Appendix~\ref{apnd:acceptance ratio}. 

The inference algorithms~\ref{alg:MHNG} and ~\ref{alg:MH-C} based on the M-H naming game can be interpreted as a process in which Agents A and B form categories from multimodal sensory-motor information and learn the cross-situations of categories in each modality and words in a word sequence proposed by others. We refer to this process as interpersonal cross-situational learning.

\subsection{Prediction as interpersonal cross-modal inference}
\label{sec:interpersonal coross-modal inference}
\begin{figure}[ht]
  \begin{center}			
  \includegraphics[scale=0.7]{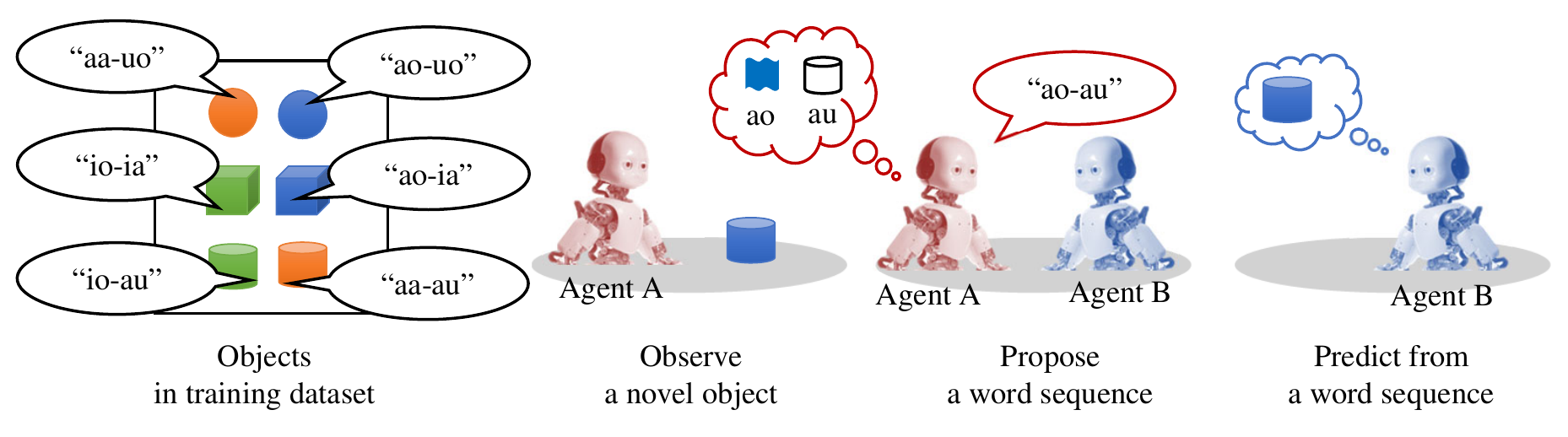}
  \caption{Example of interpersonal cross-modal inference with combinatoriality}
  \label{fig:csl_inter_cross}
  \end{center}
\end{figure}

Figure~\ref{fig:csl_inter_cross} shows an overview of the interpersonal cross-modal inference by our proposed model. The interpersonal cross-modal inference is performed between Agents A and B with the parameters inferred in the interpersonal cross-situational learning described in Section~\ref{sec:inference}. The Inter-MDM-H2H~\cite{Inter-MDM-H2H} enables agents to learn the relationship between categories and words related to objects in the training dataset. However, it was not possible to represent a novel object, which is not included in the training dataset, with combinations of learned categories and words. In the Inter-CSL-PGM, relationships between categories and words in each modality (e.g., color, object) are learned from the multimodal sensory-motor information of objects in the training dataset. For example, agents lean that ``ao'' represents blue in color modality, ``au'' represents cylinder in object modality. Therefore, if Agent A observes a blue cylinder, which is not involved in the training dataset, Agent A can propose a word sequence (e.g., ``ao-au'') based on the combinations of words associated with categories in color and object modalities, and Agent B can predict the multimodal sensory-motor information via the received word sequence and estimated categories. 

The calculation process of interpersonal cross-modal inference from Agent A to Agent B is as follows.

{\small
\begin{eqnarray}
\label{eq:cslCross_1}
z_{a,d}^{A} &\sim&
P(z_{a,d}^{A}|a_{d}^{A},{\bf{p}}_{d}^{A},A_{d}^{A},\pi_{a}^{A},{\bf{\Phi}}_{a}^{A}) \\
\label{eq:cslCross_2}
z_{p,d,A_d^A}^{A} &\sim&
P(z_{p,d,A_d^A}^{A}|p_{d,A_{d}^A}^{A},\pi_{p}^{A},{\bf{\Phi}}_{p}^{A}) \\
\label{eq:cslCross_3}
z_{o,d,A_d^A}^{A} &\sim&
P(z_{o,d,A_d^A}^{A}|o_{d,A_{d}^A}^{A},\pi_{o}^{A},{\bf{\Phi}}_{o}^{A}) \\
\label{eq:cslCross_4}
z_{c,d,A_d^A}^{A} &\sim&
P(z_{c,d,A_d^A}^{A}|c_{d,A_{d}^A}^{A},\pi_{c}^{A},{\bf{\Phi}}_{c}^{A}) .
\end{eqnarray}
}
In Equations (\ref{eq:cslCross_1})-(\ref{eq:cslCross_4}), categories ($z_{a,d}^{A}$, $z_{p,d,A_d^A}^{A}$, $z_{o, d,A_d^A}^{A}$, $z_{c,d,A_d^A}^{A}$) for a data ($d$) in each modality, i.e., action ($a$), position ($p$), object ($o$), and color ($c$), are generated using observations ($a_{d}^A$, $p_{d,A_{d}^A}^A$, $o_{d,A_{d}^A}^A$, $c_{d,A_{d}^A}^A$) for the selected object ($A_d^A$) and the parameters ($\pi^A$, $\Phi^A$) in Agnet A.

{\small
\begin{eqnarray}
\label{eq:cslCross_5}
w_{d}^{A} &\sim& P(w_{d}^{A}|z_{a,d}^{A},z_{p,d,A_d^A}^{A},z_{o,d,A_d^A}^{A},z_{c,d,A_d^A}^{A},F_d^A,{\bf{\Theta}}^{A}) .
\end{eqnarray}
}
In Equation (\ref{eq:cslCross_5}), a word sequence ($w_d^A$) for a data ($d$) is generated using the generated categories($z_{a,d}^{A}$, $z_{p,d,A_d^A}^{A}$, $z_{o, d,A_d^A}^{A}$, $z_{c,d,A_d^A}^{A}$) and parameters ($F_d^A$, ${\Theta}^{A}$) in Agent A.

{\small
\begin{eqnarray}
\label{eq:cslCross_6}
z_{a,d}^{B} &\sim&
P(z_{a,d}^{B}|w_d^A,\pi_{a}^{B},F_d^B,{\bf{\Theta}}^{B}) \\
\label{eq:cslCross_7}
z_{p,d,A_d^{B}}^{B} &\sim&
P(z_{p,d,A_d^{B}}^{B}|w_d^A,\pi_{p}^{B},F_d^B,{\bf{\Theta}}^{B}) \\
\label{eq:cslCross_8}
z_{o,d,A_d^{B}}^{B} &\sim&
P(z_{o,d,A_d^{B}}^{B}|w_d^A,\pi_{o}^{B},F_d^B,{\bf{\Theta}}^{B}) \\
\label{eq:cslCross_9}
z_{c,d,A_d^{B}}^{B} &\sim&
P(z_{c,d,A_d^{B}}^{B}|w_d^A,\pi_{c}^{B},F_d^B,{\bf{\Theta}}^{B}).
\end{eqnarray}
}
In Equations (\ref{eq:cslCross_6})-(\ref{eq:cslCross_9}), categories ($z_{a,d}^{B}$, $z_{p,d,A_d^B}^{B}$, $z_{o,d,A_d^B}^{B}$, $z_{c,d,A_d^B}^{B}$) for a data ($d$) in Agent B are generated using a word ($w_d^A$) proposed from Agent A and parameters ($\pi^{B}$, ${\Theta}^{B}$, $F_d^B$) in Agent B.

{\small
\begin{eqnarray}
\label{eq:cslCross_10}
\hat{a}_{d}^{B} &\sim&
P(\hat{a}_{d}^{B}|z_{a,d}^{B},{\bf{\Phi}}_{a}^{B}) \\
\label{eq:cslCross_11}
\hat{p}_{dA_d^B}^{B} &\sim&
P(\hat{p}_{dA_d^B}^{B}|z_{p,dA_d^B}^{B},{\bf{\Phi}}_{p}^{B}) \\
\label{eq:cslCross_12}
\hat{o}_{dA_d^B}^{B} &\sim&
P(\hat{o}_{dA_d^B}^{B}|z_{o,dA_d^B}^{B},{\bf{\Phi}}_{o}^{B}) \\
\label{eq:cslCross_13}
\hat{c}_{dA_d^B}^{B} &\sim&
P(\hat{c}_{dA_d^B}^{B}|z_{c,dA_d^B}^{B},{\bf{\Phi}}_{c}^{B}).
\end{eqnarray}
}
In Equations (\ref{eq:cslCross_10})-(\ref{eq:cslCross_13}), predicted observations ($\hat{a}_{d}^{B}$, $\hat{p}_{dA_d^B}^{B}$, $\hat{o}_{dA_d^B}^{B}$, $\hat{c}_{dA_d^B}^{B}$) for each modality in Agent B are generated using categories ($z_{a,d}^{B}$, $z_{p,d,A_d^B}^{B}$, $z_{o,d,A_d^B}^{B}$, $z_{c,d,A_d^B}^{B}$) and parameters ($\Phi^B$)

In the experiment of the interpersonal cross-modal inference described in Section~\ref{sec:eval_interpersonal cross-modal inference}, $F_d$ is fixed in the order of {action, position, object, color} to generate word sequences.

\end{CJK}

\section{Experiment}
\label{sec:experiment}
\begin{CJK}{UTF8}{min}
\subsection{Overview}

Two experiments were conducted using two humanoid robots in a simulation environment to clarify whether lexical knowledge with combinatoriality emerges between agents in our proposed model.
The first experiment is interpersonal cross-situational learning, in which the two humanoid robots take turns playing a naming game about situations on a table. In the experiment, the robots learn categories and words related to situations comprising action, position, object, and color, e.g., touching a red cube on the left side. In this experiment, the emergence of lexical knowledge with combinatoriality is evaluated based on several metrics.
The second experiment is the interpersonal cross-modal inference that was performed using the parameters obtained in interpersonal cross-situational learning. In the interpersonal cross-modal inference, Agent A proposes a word sequence for a novel situation, e.g., grasping a green cube on the right side, and Agent B predicts the sensory-motor information of the novel situation from the word sequence proposed by Agent A.
If lexical knowledge with combinatoriality emerges, Agent B can predict the sensory-motor information of novel situations by combining known words and categories in each modality.

\subsection{Conditions}
\label{sec:conditions}

\subsubsection{Simulation environment}
\label{sec:env}

\begin{figure}[ht]
  \begin{center}			
  \includegraphics[scale=1.0]{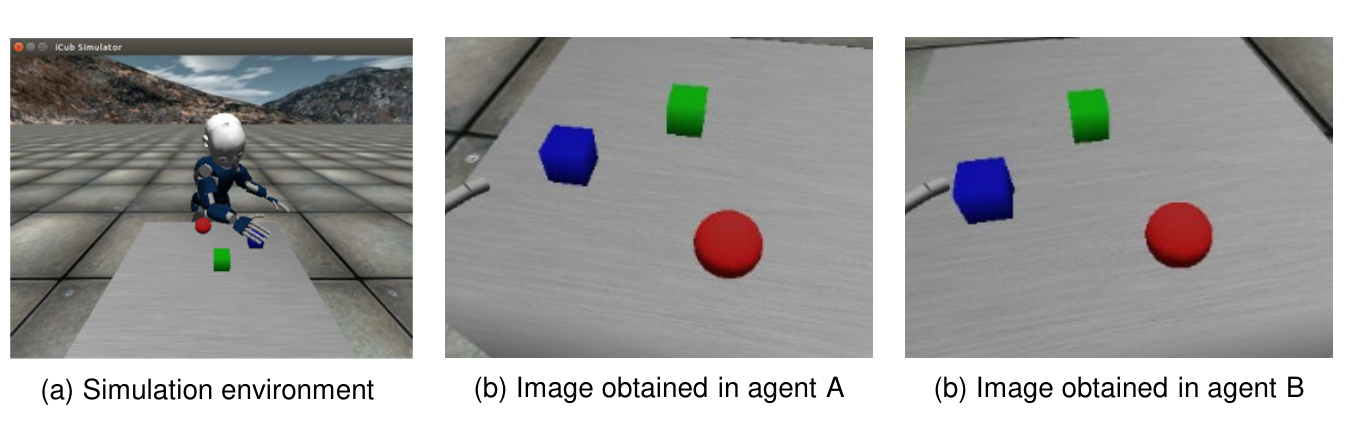}
  \caption{Experimental environment: (a) is a simulation environment based on iCub simulator, (b) and (c) show images obtained in the viewpoints of Agents A and B.}
  \label{fig:iCubsim_image}
  \end{center}
\end{figure}

The iCub simulator~\cite{tikhanoff2008} was used as a simulation environment for the experiment. 
Figure~\ref{fig:iCubsim_image} (a) shows an image captured in the simulation environment. 
In the iCub simulator, an agent is placed at the center of the environment, and objects of several shapes can be placed freely. The agent can also grab and touch objects.
In the experiment, a table was placed in front of the iCub and several objects were placed on the table.
We needed to set up two iCubs as Agents A and B to obtain their respective observations.
However, the iCub simulator had the limitation that two agents cannot appear simultaneously, and the agent's coordinates were fixed. 
Therefore, we created different observations of Agents A and B by moving the table and objects placed in front of the iCub in a pseudo manner.
This created a pseudo situation where Agents A and B were positioned next to each other facing the table.
As shown in Figure~\ref{fig:iCubsim_image} (b) and (c), Agents A and B acquire visual observations of objects from different viewpoints. The sensory-motor information comprises four modalities, i.e. action, position, object, and color. The multimodal observations were acquired in the same way as the study of CSL-PGM~\cite{CSL}.

\subsubsection{Dataset of multimodal sensory-motor information}
\label{sec:dataset}
The observations of action modality were acquired when the agent acted on a selected object.
Specifically, the body posture and relative coordinates between the right hand and the object were acquired as observations of the action modality after the action (i.e., grasping, touching, reaching out).
The details of observation for the action modality were the position of the right hand relative to the object (three dimensions), the finger openings (one dimension), the joint angles of the head, right arm, and torso (six, 16, and three dimensions) in 29 dimensions. The data were normalized to [0, 1] in each dimension.
The observation of the position modality was the coordinate data (two dimensions) of the selected object on the plane of the table in the simulation environment.
The observation of object modality was obtained from the viewpoint image as shown in Figure\ref{fig:iCubsim_image} (b) and (c). Object features were extracted from the image using CaffeNet~\cite{caffe}, a deep learning framework for convolutional neural networks (CNN)~\cite{imagenet}. The features extracted from the fc6 layer (4096 dimensions) of CaffeNet were reduced to 30 dimensions by principal component analysis (PCA) and used as observations for object modality.
The observations of color modality were obtained from the viewpoint image. 
The RGB data extracted from the image was vector quantized to ten dimensions using k-means, and the normalized version was used as the observation for the color modality.
Forty data types comprising these four modalities were prepared as observations in the experiment.
The dimension of the data and the process of feature extraction in the experiment followed the study of CSL-PGM~\cite{CSL}.

In the experiment, out of 40 observations in the dataset, 30 were used as training data for interpersonal cross-situation learning, and 10 were used as test data for interpersonal cross-modal inference. Using the parameters learned on the training data, the observations of Agent A on the test data were used as input to predict the observations of Agent B.

\subsubsection{Settings for baseline and proposed models}
\label{sec:setting}
For comparison in experiments, Inter-MDM-H2H-G was used, where the output distribution of observations in Inter-MDM-H2H~\cite{Inter-MDM-H2H} was changed to a multivariate Gaussian distribution. Futhermore, to align the outputs of the proposed model with that of the comparison model, the output distribution of observations in Inter-MDM-H2H, which was originally defined as a multinomial distribution, was changed to the same multivariate Gaussian distribution as the proposed model. The generation process and inference of Inter-MDM-H2H-G are described in Appendix~\ref{apnd:details of inter-MDM-H2H-G}. 
The hyperparameters of the Inter-CSL-PGM and the Inter-MDM-H2H-G were set as follows: $\alpha=1.0$, $\gamma=0.1$, $\beta=[m_0, \kappa_0, V_0, \nu_0]$, $\kappa_0=0.001$, $m_0=O_{x_{dim}}$, $V_0=diag(0.01,0.01)$, and $\nu_0=x_{dim}+2$, where the number of dimensions for each modality $x$ is denoted as $x_{dim}$ and the zero vector in $x_{dim}$ dimensions is denoted as $O_{x_{dim}}$. The number of categories ($L$) was set to $40$. The number of categories in each modality ($K$) was set to $10$. The number of data ($D$) was set to $40$. The number of types of words in a dictionary was set to $13$ as the dimension of $\theta$. A word dictionary consisting of 13 types of words (i.e., “aa,” “ai,” “au,” “ae,” “ao,” “ia,” “ii,” “iu,” “ie,” “io,” “ua,” “ui,” “uu”) was used.

\subsection{Evaluation in interpersonal cross-situational learning}
\label{sec:exp_learning}
\subsubsection{Procedure}
\label{sec:procedure1}
\begin{figure}
  \begin{center}			
  \includegraphics[scale=0.85]{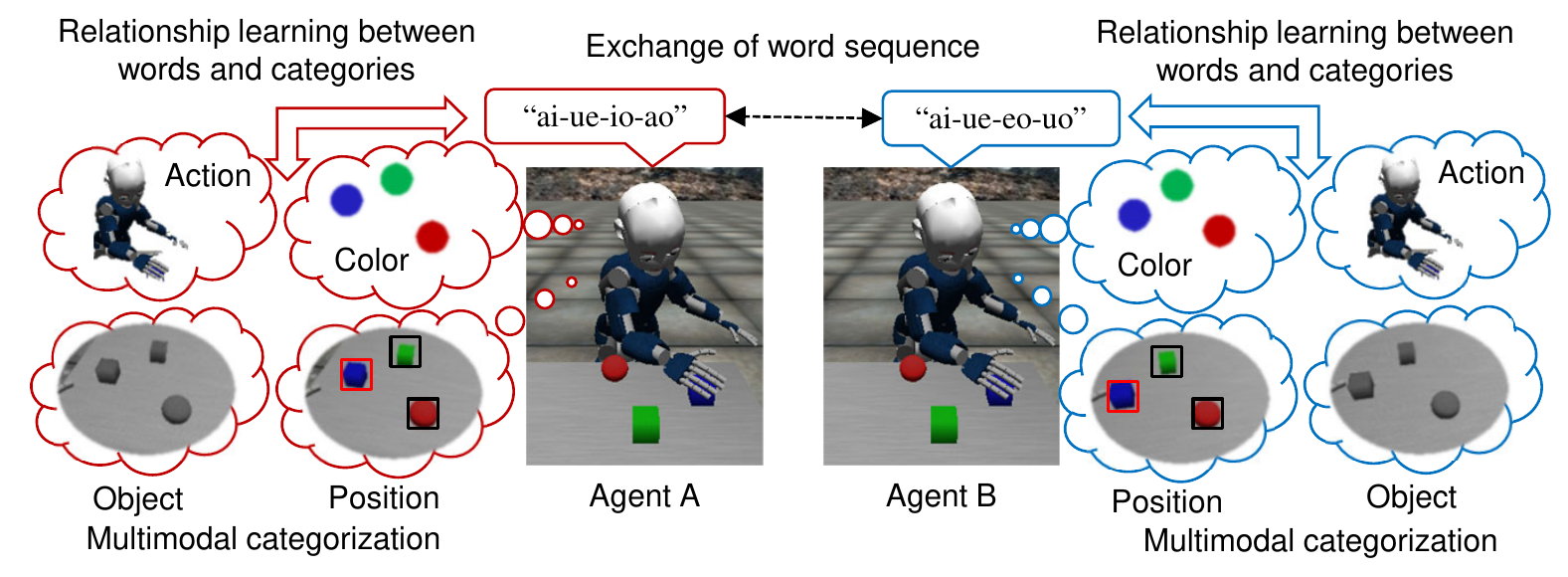}
  \caption{Overview of interpersonal cross-situational learning}
  \label{fig:condition1}
  \end{center}
\end{figure}

In the experiment, two humanoid robots took turns playing a naming game about objects on a table.
As shown in Figure~\ref{fig:condition1}, the robots learned categories in each modality and words related to objects by exchanging a word sequence.
Interpersonal cross-situational learning was performed using the following procedure.
\begin{enumerate}
 \item Multiple objects are placed on a table in front of Agents A and B.
 \item Agent A obtains multimodal information (i.e., position, object, and color information) about all objects on the table.
 \item Agent A selects one of the objects on the table.
 \item Agent A selects an action for the selected object from grasping, touching, or reaching out. At this time, the agent obtains action information.
 \item Agent A proposes a word sequence for the selected object and the action.
 \item Agent B accepts or rejects the proposed word sequence and updates its own parameters.
\end{enumerate}
This sequence of steps was alternated between Agents A and B. 
The object and action selected by Agents A and B were assumed to be the same. 
In this iteration, the agents learned relationships between words in word sequences and categories in each modality.

\subsubsection{Evaluation criteria and baseline models}
\label{sec:criteria1}
We verified our proposed model in the following items with evaluation criteria to confirm the emergence of lexical knowledge with combinatoriality among agents.

\begin{enumerate}
    \item To verify whether the words associated with the categories in specific modalities had been learned, we analyzed the distributions ($\theta$) to generate words from the category index in each modality and evaluate the relationship between words and modalities based on the normalized mutual information (NMI) compared with the baseline model.
    \item To verify whether categories that capture the situational elements had been formed in each modality, we analyzed categorized data in each modality and evaluate the categorization accuracy based on the adjusted Rand index (ARI) between estimated categories and ground truth.
    \item To verify whether words associated with categories in each modality and word sequences representing situations are shared among agents, we evaluated similarities in words and word sequences based on Kappa coefficient ($\kappa$) and estimation accuracy rate (EAR) compared with baseline models.
\end{enumerate}

As evaluation criteria, the NMI was used to evaluate the relationship between words and modalities, the ARI~\cite{Hubert85} was used to evaluate the categorization accuracy in each modality, and the Kappa coefficient ($\kappa$)~\cite{Cohen60} was used to evaluate the agreement in word usage between agents. The EAR was used to evaluate the error in word sequence usage between agents. 

The NMI was calculated using the following equation:
{\small
\begin{eqnarray}
    \label{eq:nmi}
    {\rm NMI}(w;F) &=& \frac{I(w;F)}{\sqrt{H(w)H(F)}},
\end{eqnarray}
}
where $w$ is a random variable indicating the uttered words, $F$ is a random variable indicating the modalities corresponding to $w$. $I(\cdot)$ is the mutual information, $H(\cdot)$ is the entropy.

The ARI was calculated using the following equation:
{\small
\begin{eqnarray}
\label{eq:ari}
\rm ARI &=& \frac{{\rm RI-Expected\:RI}}{{\rm \max (RI)-Expected\:RI}},
\end{eqnarray}
}
where RI is the Rand Index.

The Kappa coefficient ($\kappa$) was calculated using the following equation:
{\small
\begin{eqnarray}
\label{eq:kp}
\kappa &=& \frac{C_o - C_e}{1 - C_e},
\end{eqnarray}
}
where $C_o$ is the coincidence rate of signs between agents, and $C_e$ is the coincidence rate of signs between agents by random chance. The $\kappa$ value is judged as follows: ($0.81-1.00$) as almost perfect agreement, ($0.61-0.80$) as substantial agreement, ($0.41-0.60$) as moderate agreement, ($0.21-0.40$) as fair agreement, ($0.00-0.20$) as slight agreement, and ($\kappa<0.0$) as no agreement~\citep{Criteria}.

The EAR was calculated using the following equation:
{\small
\begin{eqnarray}
\label{eq:ari}
\rm EAR &=& 1- \frac{{\rm \text{The number of estimation errors}}}{{\rm \text{The number of all of uttered words}}}.
\end{eqnarray}
}
Although EAR is a metric calculated using words spoken by a human tutor as the ground truth in the study of CSL-PGM~\cite{CSL}, here, it was calculated using words generated by Agent A as the ground truth.

To evaluate our proposed model based on these metrics, we used  the Inter-CSL-PGM (w/o MEC) and the Inter-CSL-PGM (No communication) as the baseline models. 
The Inter-CSL-PGM (w/o MEC) is a variation of the Inter-CSL-PGM with the re-scaling term in Equations~\ref{eq:rescaling} set to one. This can be interpreted as a model that does not assume the mutual exclusivity bias in which infants assign novel words to unknown objects in lexical acquisition. The Inter-CSL-PGM (No communication) is a variation of the Inter-CSL-PGM with the acceptance rate ($\alpha$) in Algorithm~\ref{alg:MH-C} set to zero. As it rejects all utterances from another agent, it can be interpreted as a model that learns independently without engaging in communication.

\subsubsection{Experimental Result: Relationship between words and modalities}

\label{sec:result1}
\begin{figure}
  \begin{center}			
  \includegraphics[scale=0.75]{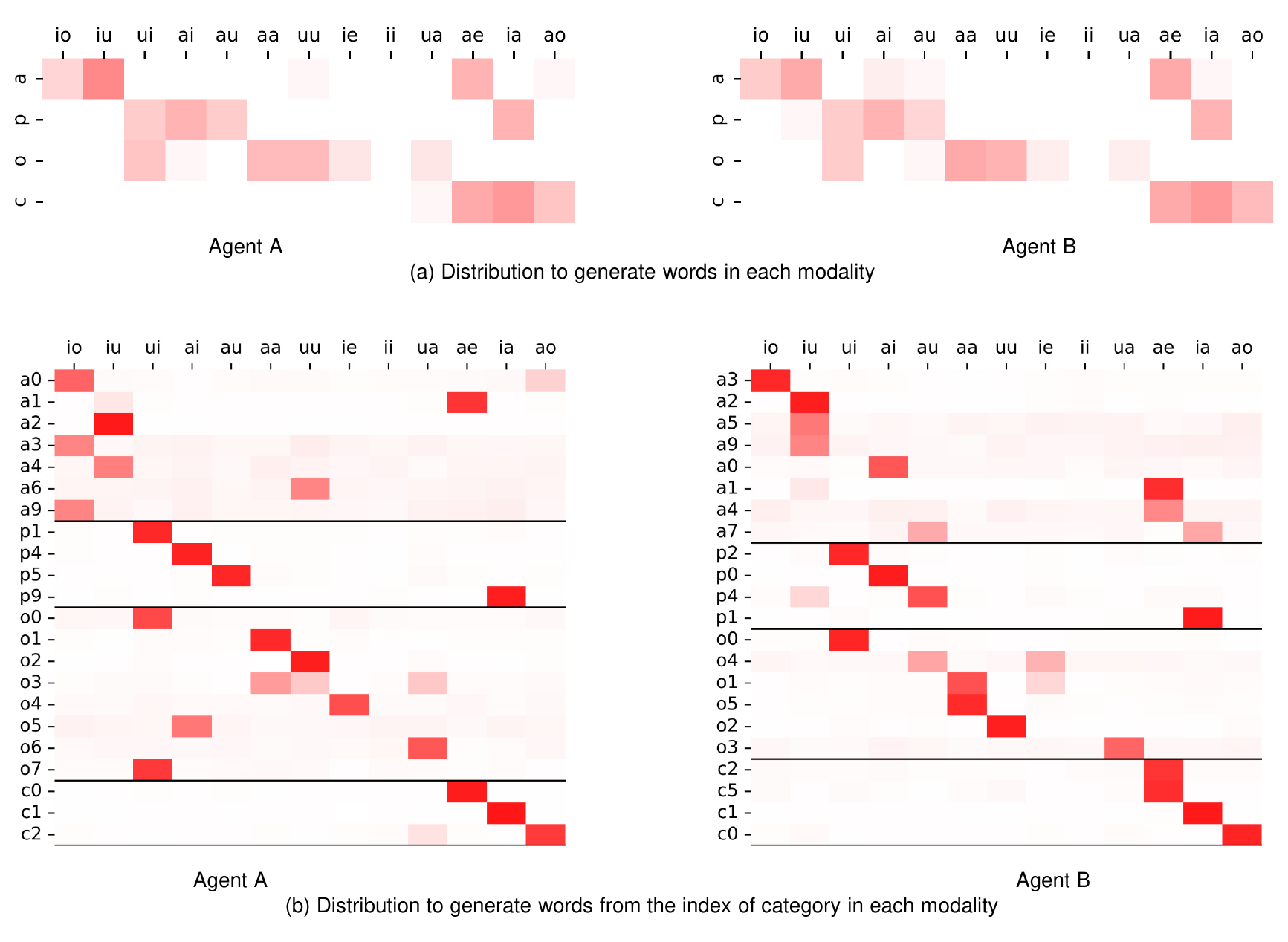}
  \caption{The word probability ($\theta$) estimated in the experiment of interpersonal cross-situational learning. (a) and (b) show the probability ($\theta$) in each modality and category, respectively. A, p, o, and c show action, position, object, and color modalities in each graph. The darker shade shows the probability ($\theta$). The rows and columns show modalities and words in (a), and categories and words in (b).}
  \label{fig:exp_cls}
  \end{center}
\end{figure}
To verify whether the words associated with the categories in specific modalities had been acquired in the agents through interpersonal cross-situational learning, we analyzed the distributions of words ($\theta$) in each modality (i.e. action, position, object, and color).
Figure~\ref{fig:exp_cls} shows the distribution ($\theta$) to generate words from the index of categories in each modality. (a) and (b) show the probability of words in each modality and category, respectively. A higher probability value was represented by a darker shade. If the words associated with modality could be learned correctly, the probabilities of specified words became higher for each modality (i.e., action, position, object, and color) in the graphs of (a). For example, the color modality shows higher probabilities for words (i.e., “ae,” “ia,” and “ao”) in Agents A and B. The other modalities are also the same. The result suggests that words associated with modalities are shared among agents. However, some words (e.g., “ui,” “ae,” and “ia”) show higher probabilities in multiple modalities. For example, in the graph of Agent A in (b), the word (“ae”) has higher probabilities in the categories of a1 and c0. This is presumably because these two categories were only observed simultaneously in the training dataset and were not assigned to different words. For the color and position categories in the graphs of (b), a tendency for different words to have higher probabilities in each category can be confirmed. 
\begin{table}
\centering
\caption{The NMIs between the variables of modality and word in each agent. The NMI takes a value between 0 and 1, where 0 indicates no relationship and values closer to 1 indicate a stronger relationship between two variables. Bold and underscore indicate the highest values in each agent.}
\label{tab:result-nmi}
\scalebox{0.85}{
\begin{tabular}{l cc cc}
\hline
 & \multicolumn{2}{c}{Agent A} & \multicolumn{2}{c}{Agent B} \\
\multicolumn{1}{c}{Model} & Mean & SD & Mean & SD \\ \hline
Inter-CSL-PGM (Proposed) & \underline{\textbf{0.583}} & 0.054 & \underline{\textbf{0.572}} & 0.046 \\
Inter-CSL-PGM (w/o MEC) & 0.278 & 0.127 & 0.278 & 0.126 \\
Inter-CSL-PGM (No communication) & 0.248 & 0.039 & 0.223 & 0.034 \\ \hline
\end{tabular}
}
\end{table}

To quantitatively evaluate the relationship between words and modalities, we calculated the NMIs from the distributions $\theta$ in Figure \ref{fig:exp_cls}(a). Table~\ref{tab:result-nmi} shows NMIs in the proposed and baseline models. Comparing the NMIs, the proposed model has values more than twice those of the baseline models in Agents A and B, indicating a high relationship between words and modalities. 
In Inte-CSL-PGM (w/o MEC), since there is no bias in assigning novel words to new categories, the probability to generate a certain word across multiple modalities increased, resulting in lower NMIs.
In Inter-CSL-PGM (No communication), the learning of word distributions associated with categories did not progress due to the absence of word proposals from another agent, resulting in lower NMIs.
These results demonstrated that the proposed model enables agents to acquire a lexicon related to specific modalities. It is expected that agents can represent novel situations with word sequences by combining words related to each modality, e.g., the combination of “green” in color modality and “cup” in object modality.
Furthermore, it was suggested that the mutual exclusivity constraint, which assigns novel words to unknown situational elements, plays a crucial role in interpersonal cross-situational learning for the emergence of lexical knowledge with combinatoriality.

\begin{figure}
  \begin{center}			
  \includegraphics[scale=0.75]{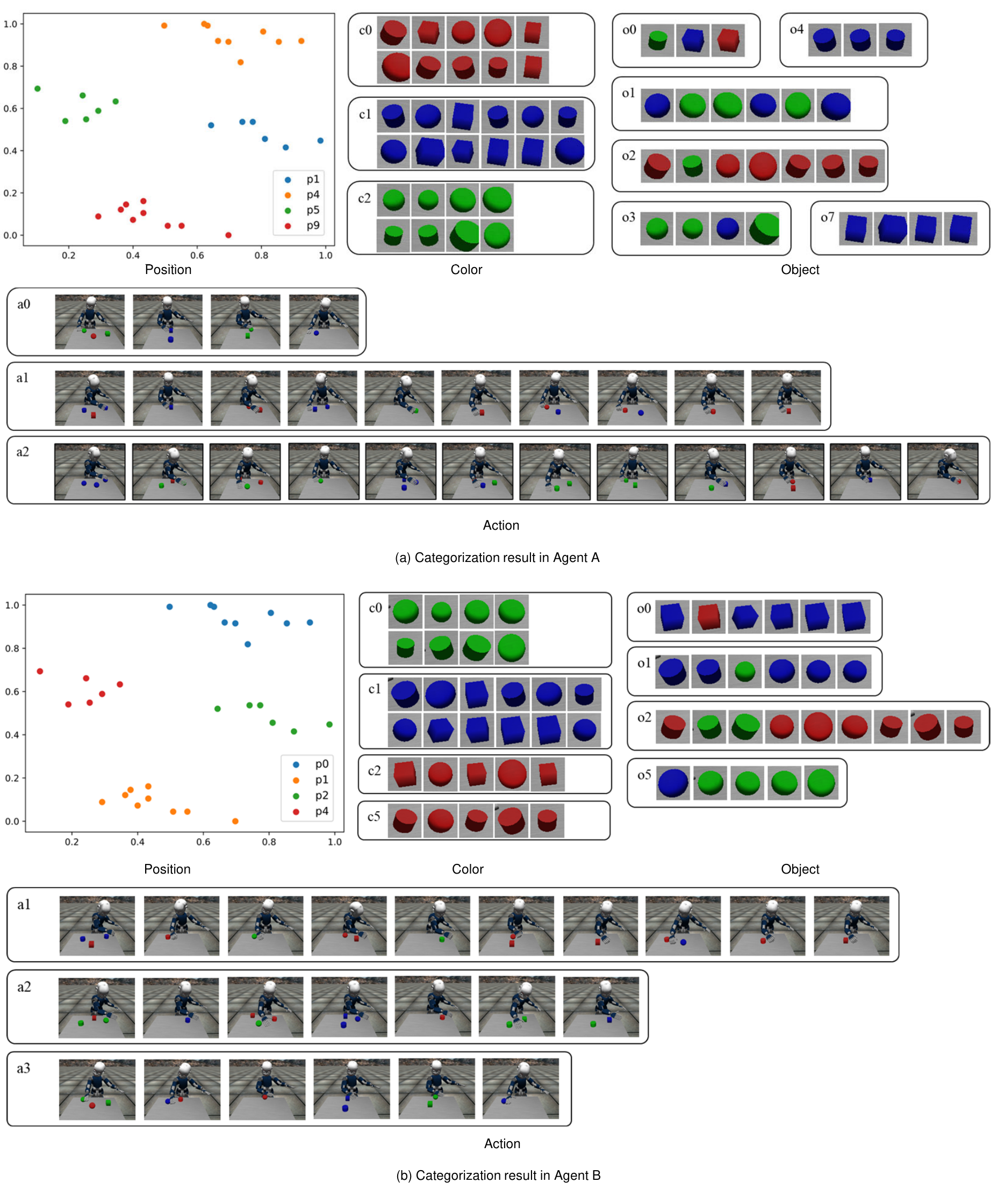}
  \caption{Categorization results in the experiment of interpersonal cross-situational learning. (a) and (b) show the categorization results of Agents A and B, respectively. The number shows the index of category in each modality (i.e. p: position, c: color, o: object, and a: action). Only categories with three or more assigned data points are shown.}
  \label{fig:exp_category}
  \end{center}
\end{figure}

\subsubsection{Experimental Result: Categorization accuracy and word sharing}

To verify whether the formed categories capture situational elements in each modality, we analyzed the results of categorization. 
Figures~\ref{fig:exp_category} (a) and (b) show the categorization results in Agents A and B, respectively. 
The left graph shows the positions of objects in a two-dimensional coordinate system on the table, and the color of the dot shows the categorization results in the position modality.
In Figure~\ref{fig:exp_category} (a), Agent A categorized the data on the right side of the table into a position category (p1), and the category is associated with the word (“ui”) in Figure~\ref{fig:exp_cls} (b). 
In Figure~\ref{fig:exp_category} (b), Agent B categorized the data on the right side of the table to the position category (p2), and the category is also associated with the word (“ui”) in Figure~\ref{fig:exp_cls} (b). 
Comparing the position categories of Agents A and B shows that the same words were learned to represent the distribution of closed positions.
In the color category of Agent A, c0 is associated with the word (“ae”).
The category associated with the word (“ae”) in agent B is c2 and c5 in the color modality.
It is shown that c0 in Agent A and c2 and c5 in Agent B have been learned and shared as a category meaning red color.
Similar results were obtained in other color categories.
Agents A and B correctly obtained color categories and shared the words associated with categories.
In the categorization results in object modality, categories that capture the objects were formed. 
In Figure~\ref{fig:exp_category} (a), cups were categorized to o4, and boxes were categorized to o7.
However, categories with a mixture of different objects, such as o0, have also been formed.
This may be due to the simplicity of the three-dimensional shape of the objects, which made it difficult for the agent to distinguish between objects with different shapes. 
In the categorization results in action modality, categories related to hand shapes were formed. 
In Figure~\ref{fig:exp_category} (a), the reaching action with an open hand was categorized as a0, and the grasp action with a closed hand was categorized as a1. 
However, a category with a mixture of reaching and touch actions, such as a2, was also formed.
This may be because the agent's posture differs depending on the object's position.
Because of these categorization problems, it is thought that in the object and action modalities, one word was associated with multiple categories within the modality in Figure~\ref{fig:exp_cls} (b).
Although not perfect for the action and object categories, the experimental results demonstrated that words corresponding to the situational elements associated with the modality were learned and shared among the agents.

\begin{table}
 \centering
 \caption{The evaluation of categorization accuracy based on ARI, the similarities of words based on $\kappa$, and the errors of word sequences based on EAR. The a, p, o, and c show action, position, object, and color modalities, respectively. Bold and underscore indicate the highest values, and bold indicates the second-highest values.}
 \scalebox{0.67}{
 \begin{tabular}{l l c c c c c c c c c}
 \hline
 Models & Trials & ARI\_a & ARI\_p & ARI\_o & ARI\_c & $\kappa$\_a & $\kappa$\_p & $\kappa$\_o & $\kappa$\_c & EAR \\ \hline 
 Inter-CSL-PGM (Proposed) & 30 & 0.309 & \textbf{0.794} & \textbf{0.286} & \textbf{0.949} & \textbf{\underline{0.677}} & \textbf{\underline{0.883}} & \textbf{\underline{0.671}} & \textbf{0.864} & \textbf{\underline{0.792}} \\
 Inter-CSL-PGM (w/o MEC) & 30 & \textbf{0.310} & 0.651 & 0.250 & \textbf{\underline{0.961}} & \textbf{0.569} & \textbf{0.633} & \textbf{0.639} & \textbf{\underline{0.919}} & \textbf{0.431}\\
 Inter-CSL-PGM (No communication) & 30 & \textbf{\underline{0.358}} & \textbf{\underline{0.831}} & \textbf{\underline{0.306}} & 0.930 & 0.017 & 0.011 & 0.002 & 0.027 & 0.243 \\
 \hline \hline
 CSL-PGM & 20 & 0.300 & 0.606 & 0.408 & 0.782 & - & - & - & - & 0.970 \\ 
 CSL-PGM & 40 & 0.375 & 0.540 & 0.366 & 0.870 & - & - & - & - & 0.989 \\ 
  \hline
 \end{tabular}
 }
 \label{tab:cls_result}
\end{table}

We quantitatively evaluated the categorization accuracy in each agent and the sharing of words and word sequences among agents. Table~\ref{tab:cls_result} shows the ARI between estimated categories and ground truth, the Kappa coefficient ($\kappa$) between words among agents, and the EAR of word sequences among agents.  
The first, second, and third rows are the results of the proposed model and the baseline models. The fourth and fifth rows show the experimental results of CSL-PGM conducted in the study~\cite{CSL} using the same data set and hyperparameters as reference data. Note that the experimental conditions of CSL-PGM are different because the human tutor provided the instruction sentences.
First, in terms of the ARI, which represents categorization accuracy, no significant differences were confirmed between our proposed model and the baseline models. In these models, the ARI was high for position and color modalities, while it was low for action and object modalities. These results are consistent with the analysis presented in the previous paragraph, and the same trend can be observed in the CSL-PGM. This indicates that valid categorization was achieved in all models in this experimental setting.
Next, in terms of $\kappa$, which represents the degree of agreement in words among agents, the proposed model demonstrated higher values compared to the Inter-CSL-PGM(No communication). Based on the $\kappa$ metric, action and object exhibit substantial agreement, while position and color exhibit perfect agreement. Furthermore, the proposed model also showed higher values for the EAR, which represents the error in word sequences among agents, compared to the baseline models, approaching the values of the CSL-PGM as the top line.
The proposed model enabled sharing of words and word sequences among agents while maintaining categorization accuracy for each agent.
The experimental results of interpersonal cross-situational learning demonstrated that the M-H naming game enables the emergence of not only categorical signs but also lexical knowledge with combinatoriality dependent on modalities.

\subsection{Evaluation in the interpersonal cross-modal inference}
\label{sec:eval_interpersonal cross-modal inference}
\subsubsection{Procedure}
\label{sec:procedure2}
\begin{figure}
  \begin{center}			
  \includegraphics[scale=0.85]{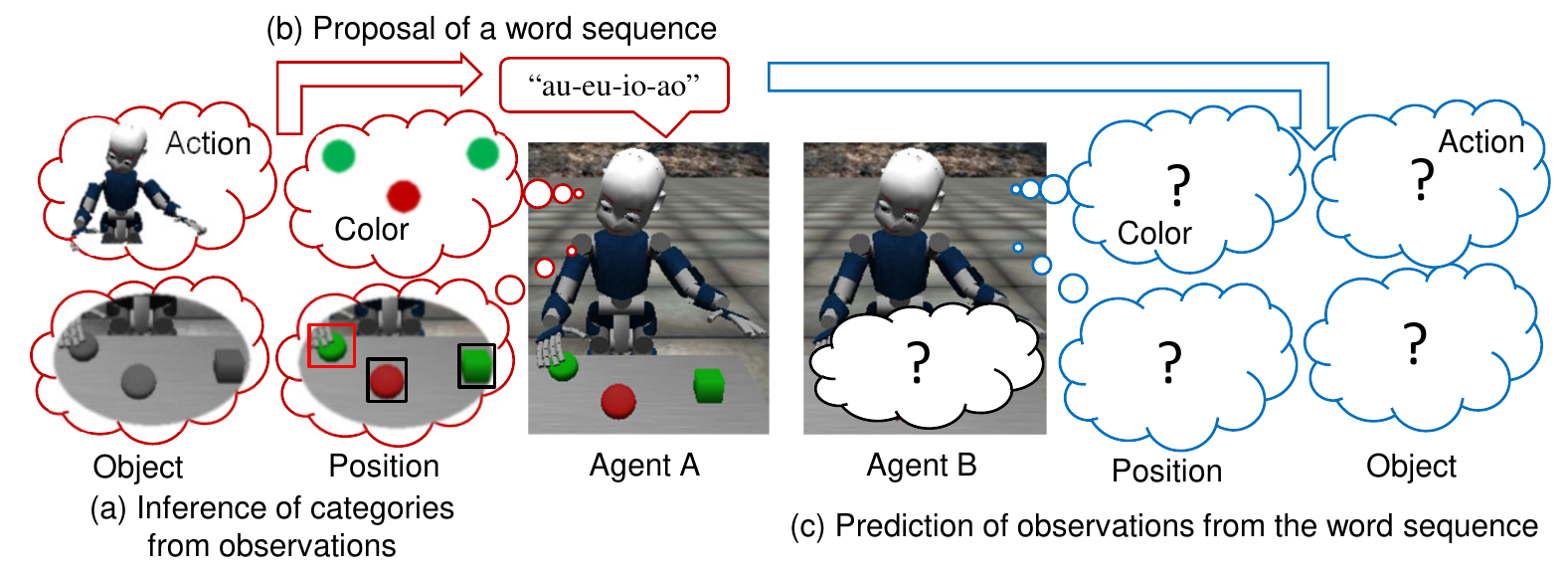}
  \caption{Overview of interpersonal cross-modal inference}
  \label{fig:condition2}
  \end{center}
\end{figure}
To evaluate the generalization performance of lexical knowledge obtained in interpersonal cross-situational learning for novel situations, we conducted an experiment on interpersonal cross-modal inference as shown in Figure~\ref{fig:condition2}.
The experimental procedure of interpersonal cross-modal inference was described as follows:

\begin{enumerate}[(a)]
\item Agent A infers categories from observations of a novel situation.
\item Agent A proposes a word sequence based on the inferred categories.
\item Agent B predicts observations from the proposed word sequence.
\item The roles of Agents A and B are alternated and repeated.
\end{enumerate}

If the categories capturing situational elements and the lexicons related to specific modalities are shared among agents, it is expected that Agents A and B can predict the observations of novel situations with small errors through the combinations of these words.

\subsubsection{Evaluation criteria and baseline models}
\label{sec:criteria1}
As an evaluation criterion in the experiment, we used the mean squared error (MSE) between predicted sensor-motor information by the agent and the original sensor-motor information of novel situations.
The equation for MSE is shown below. 
{\small
\begin{eqnarray}
\label{eq:mse}
{\rm MSE} = \frac{1}{D}\sum_{d=1}^{D} \left(\hat{y_d} - y_d\right)^2,
\end{eqnarray}
}
where $\hat{y_d}$ denotes the predicted value, $y_d$ denotes the original value, and $D$ is the number of data.

To evaluate our proposed model, we used the Inter-CSL-PGM (w/o MEC), Inter-CSL-PGM (No communication), and Inter-MDM-H2H-G as the baseline models.
Inter-MDM-H2H-G is a model that modifies the Inter-MDM-H2H~\cite{Inter-MDM-H2H}, which is a model of the M-H naming game, by changing the distribution for generating observations from a multinomial distribution to a Gaussian distribution. The detail of Inter-MDM-H2H-G is described in Appendix~\ref{apnd:details of inter-MDM-H2H-G}. In this model, multimodal information is generated from an integrated single category. Consequently, it is not possible to learn words associated with categories in specific modalities.

\subsubsection{Experimental results}

\begin{table}
 \centering
 \caption{Comparison of MSEs for our proposed model and Inter-MDM-H2H-G in interpersonal cross-modal inference. Mean and SD shows the average and standard deviation of MSEs. The smallest values in MSEs are indicated in bold underlined, and bold indicates the second-smallest values.}
 \scalebox{0.80}{
 \begin{tabular}{l c c c c c c c c}
 \hline
  & \multicolumn{2}{c}{Action} & \multicolumn{2}{c}{Position} & \multicolumn{2}{c}{Object} & \multicolumn{2}{c}{Color} \\ 
 \multicolumn{1}{c}{Algorithm} & Mean & SD & Mean & SD & Mean & SD & Mean & SD \\ \hline 
 Inter-CSL-PGM (Proposed) & \textbf{\underline{1.004}} & 0.377 & \textbf{\underline{0.012}} & 0.007 & \textbf{\underline{0.278}} & 0.063 & \textbf{\underline{0.145}} & 0.058 \\ 
 Inter-CSL-PGM (w/o MEC) & \textbf{1.328} & 0.334 & \textbf{0.024} & 0.006 & 0.387 & 0.196 & \textbf{0.478} & 0.126 \\
Inter-CSL-PGM (No communication) & 1.471 & 0.300 & 0.043 & 0.007 & 0.529 & 0.088 & 0.742 & 0.208 \\
 Inter-MDM-H2H-G & 1.383 & 0.224 & 0.031 & 0.008 & \textbf{0.382} & 0.043 & 0.612 & 0.129 \\ \hline 
 \end{tabular}
 }
 \label{tab:result1}
\end{table}

Table~\ref{tab:result1} shows a comparison of the MSEs in our proposed model and baseline models in interpersonal cross-modal inference. The MSEs are shown with mean and standard deviation (SD) in each modality.
In the result shown in Table~\ref{tab:result1}, the MSEs in our proposed model demonstrated lower values than the MSEs in baseline models for all modalities. This means our proposed model enables the agent to predict sensory-motor information closer to the original sensory-motor information compared with baseline models. 

However, it's unclear whether our proposed model can predict the sensory-motor information of novel situations based on the combinations of words learned in interpersonal cross-situational learning.
To address this, we conducted interpersonal cross-modal inference on both known situations (training dataset) and novel situations (test dataset) using the parameters estimated in interpersonal cross-situational learning. Here, we refer to the observations of the known situations included in interpersonal cross-situational learning as the training dataset, and the observations of the novel situations as the test dataset.
Generally, in the computational models, the prediction errors increase on the test dataset compared to the training dataset. However, if the novel situations in the test dataset can be represented by combining the situational elements (e.g. action, position, object, and color) in the training dataset, it is expected that the prediction errors will not significantly increase. To confirm this, we compared the MSEs in our proposed model and baseline models for the training and test datasets. Table~\ref{tab:result2} presents the MSEs in our proposed model and baseline models for the training and test datasets as the result of interpersonal cross-modal inference. The t-tests were performed on the training and test datasets in each model.

\begin{table}
 \centering
 \caption{Comparison of MSEs in our proposed model and baseline models for the training and test datasets as the results of interpersonal cross-modal inference. We show the average and standard deviation of MSEs in ten trials. (.) indicates the standard deviation. t indicates the t-test. In the t-test, **: $(p<0.01)$, *: $(p<0.05)$, n.s.: not significant, H0: MSEs in the training and test datasets are equal, H1: MSEs in test data are larger than MSEs in training data. In the training and test datasets, the smallest and second-smallest MSEs are indicated in bold underline and bold, respectively.}
 \scalebox{0.65}{
 \begin{tabular}{l c c c c c c c c c c c c}
 \hline
  & \multicolumn{3}{c}{Action} & \multicolumn{3}{c}{Position} & \multicolumn{3}{c}{Object} & \multicolumn{3}{c}{Color} \\ 
 Model & Training & Test & t & Training & Test & t & Training & Test & t & Training & Test & t \\ \hline
 Inter-CSL-PGM & \textbf{1.083} & \textbf{\underline{1.004}} & n.s. & \textbf{\underline{0.011}} & \textbf{\underline{0.012}} & n.s. & \textbf{0.362} & \textbf{\underline{0.278}} & n.s. & \textbf{\underline{0.202}} & \textbf{\underline{0.145}} & n.s. \\ 
 (Proposed) & (0.197) & (0.377) & & (0.002) & (0.007) & & (0.074) & (0.063) &  & (0.046) & (0.058) &  \\ \hline
 Inter-CSL-PGM & 1.274 & \textbf{1.328} & n.s. & \textbf{0.032} & \textbf{0.024} & n.s. & 0.404 & 0.387 & n.s & 0.494 & \textbf{0.478} & n.s \\ 
 (w/o MEC) & (0.220) & (0.334) & & (0.005) & (0.006) & & (0.046) & (0.196) & & (0.069) & (0.126) & \\ \hline
Inter-CSL-PGM & 1.440 & 1.471 & n.s. & 0.041 & 0.043 & n.s. & 0.503 & 0.529 & n.s. & 0.747 & 0.742 & n.s. \\ 
(No communication) & (0.199) & (0.300) & & (0.006) & (0.007) & & (0.045) & (0.088) & & (0.067) & (0.208) & \\ \hline
 Inter-MDM-H2H-G & \textbf{\underline{0.820}} & 1.383 & $**$ & 0.033 & 0.031 & n.s. & \textbf{\underline{0.348}} & \textbf{0.382} & $*$ & \textbf{0.440} & 0.612 & $**$ \\ 
 & (0.138) & (0.224) & & (0.006) & (0.008) & & (0.031) & (0.043) & & (0.100) & (0.129) & \\ \hline
 \end{tabular}
 }
 \label{tab:result2}
\end{table}

In Inter-MDM-H2H-G, which forms multimodal integrated categories representing situations, significant differences were found in the test dataset compared to the training dataset in three modalities (i.e. action, object, and color). The result of Inter-MDM-H2H-G suggests that the integrated categories learned in the training dataset did not have generalization performance for novel situations in the test dataset. 
In contrast, the t-test results in our proposed model indicate no significant difference for all modalities in MSEs between the training and test datasets. This suggests that our proposed model can predict the sensory-motor information of novel situations by combining words associated with specific modalities. 
Despite the no significant differences between the training and test datasets in both the w/o MEC and no communication models of Inter-CSL-PGM, the MSEs were larger compared to our proposed model. The lack of sharing words associated with categories among the agents is considered to be the cause of this result.
This result implies that our proposed model has generalization performance to novel situations in interpersonal cross-modal inference compared to baseline models. 
It is also suggested that the mutual exclusivity bias and the M-H naming game enable better predictions of novel situations in interpersonal cross-modal inference.

\subsubsection{Qualitative evaluation}
\begin{figure}
  \begin{center}			
  \includegraphics[scale=0.73]{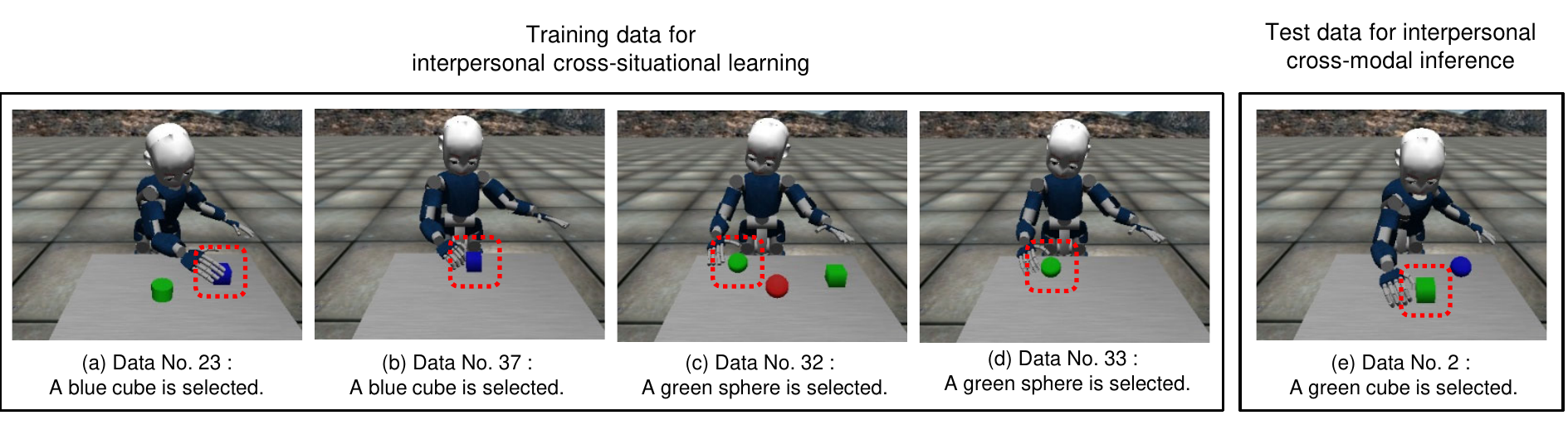}
  \caption{An example of scenes: (a) to (d) show scenes in which the robot obtains training data in interpersonal cross-situational learning, and (e) shows a scene in which the robot obtains test data in interpersonal cross-modal inference. A red rectangle with a dashed line in each scene shows a selected object.}
  \label{fig:resutl}
  \end{center}
\end{figure}
We evaluated whether a word sequence that represents a novel situation could be generated by the combinations of words learned in interpersonal cross-situational learning. Figures~\ref{fig:resutl} (a)-(d) show the scenes for obtaining training data in interpersonal cross-situational learning. A blue box was selected as an attention object in Scenes (a) and (b). In Scenes (c) and (d), a green ball was selected. Scene (e) shows the scene for obtaining the test data in interpersonal cross-modal inference. In Scene (e), the robot observed a green box, which was not included in the training data, as a novel situation. 
We confirmed whether a word sequence that represents the green box could be generated by the combination of words representing the categories in color and object modalities learned in interpersonal cross-situational learning.

\begin{table}
 \centering
 \caption{Results of the generation of word sequences from the observations in Scenes (a)-(e) by the proposed model. Words representing a novel scene in (e) are indicated with bold underline.}
 \scalebox{0.85}{
 \begin{tabular}{c c c c c c}
 \hline
 Modality & Scene (a) & Scene (b) & Scene (c) & Scene (d) & Scene (e)\\ \hline
 
 Action   & iu & ae & io & iu & ae \\ 
 Position & ao & ui & au & ai & ao \\
 Object   & \textbf{\underline{ui}} & \textbf{\underline{ui}} & aa & aa & \textbf{\underline{ui}} \\
 Color    & ia & ia & \textbf{\underline{ao}} & \textbf{\underline{ao}} & \textbf{\underline{ao}} \\ \hline
 \end{tabular}
 }
 \label{tab:result3}
\end{table}
Table~\ref{tab:result3} shows the generated word sequences from the observations of each scene using the parameters learned by our proposed model. In Scenes (a) and (b), the word “ui” was generated in  object modality, and the word “ia” was generated in color modality. It is inferred that “ui” and “ia” were learned to represent the box and blue, respectively. In Scenes (c) and (d), the word “aa” was generated in object modality, and the word “ao” was generated in color modality. It is inferred that “aa” and “ao” were learned to represent the ball and green, respectively. It is confirmed that the green box, a novel situation, can be represented by the combinations of the word “ui” representing the box in the object modality, and “ao” representing the green in the color modality. However, it is unclear from Table~\ref{tab:result3} whether words such as “ao” and “ui” were learned to represent color and object categories related to green color and box object. Therefore, we analyzed the categories associated with “ao” and “ui”.

\begin{figure}
    \begin{tabular}{c}
      \begin{minipage}[t]{0.9\hsize}
        \centering
        \includegraphics[keepaspectratio, scale=0.67]{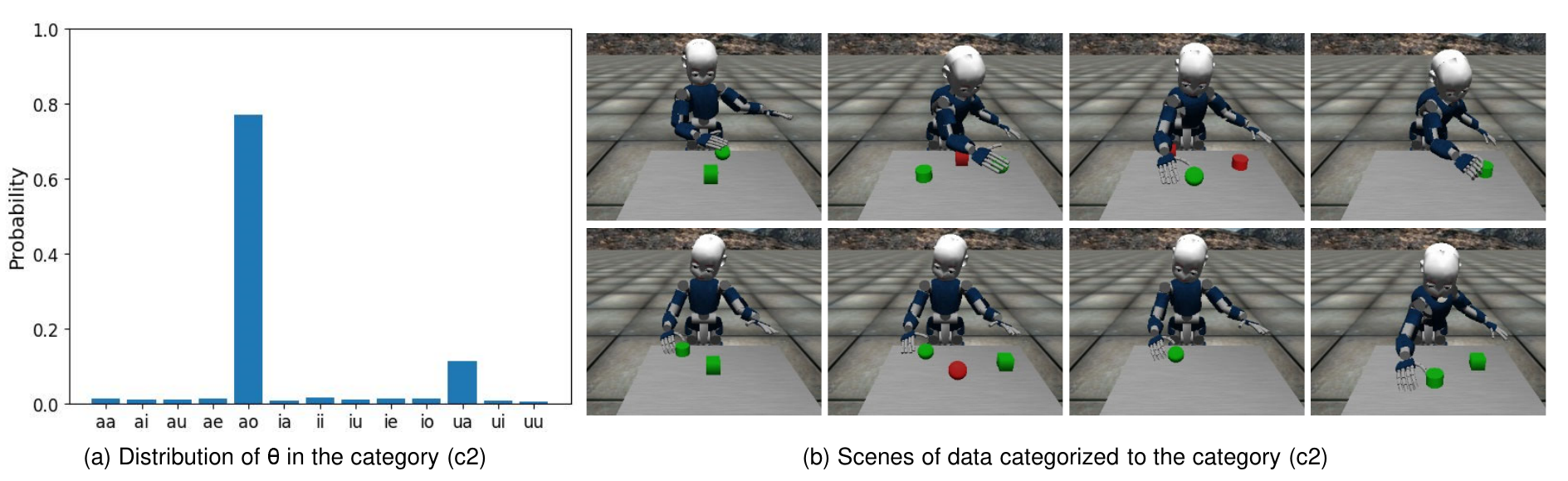}
        \caption{Category (c2) in color modality associated with “ao” and scenes of data assigned to the category}
        \label{fig:c2}
      \end{minipage} \\
   
      \begin{minipage}[t]{0.9\hsize}
        \centering
        \includegraphics[keepaspectratio, scale=0.67]{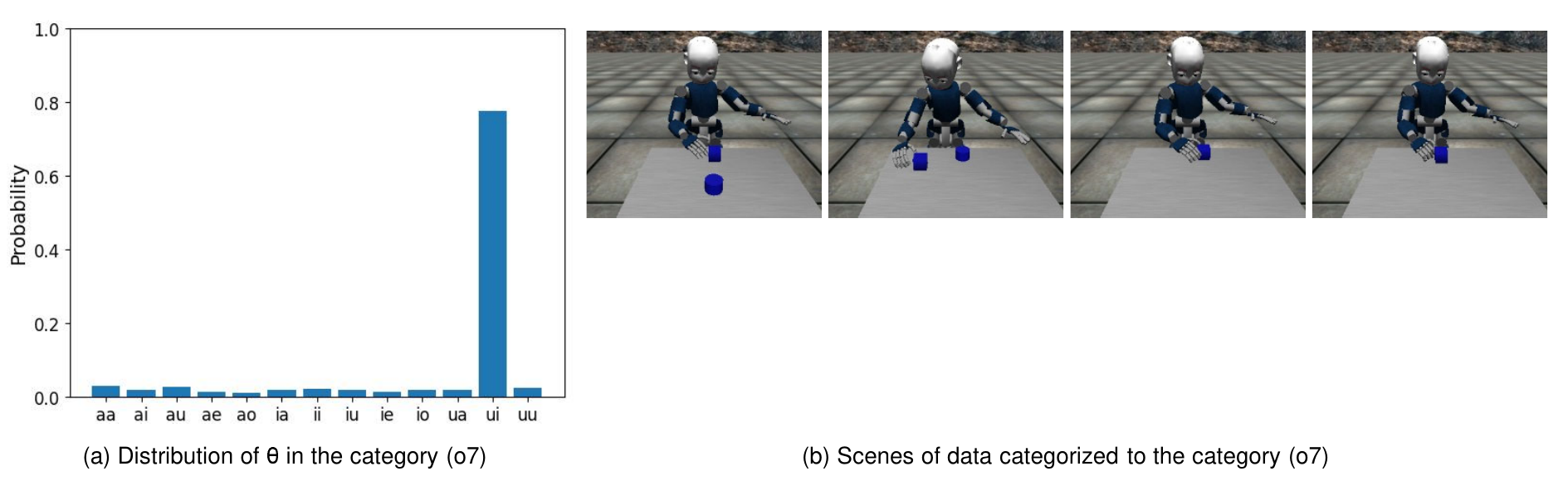}
        \caption{Category (o7) in object modality associated with “ui” and scenes of data assigned to the category}
        \label{fig:o7}
      \end{minipage}
    \end{tabular}
\end{figure}
Figure~\ref{fig:c2} shows Category (c2) in color modality associated with “ao” and the scenes of data assigned to the category. According to the distribution of $\theta$ in (a), “ao” has a high probability of generating words in Category (c2). This implies that Category (c2) was associated with “ao”.
(b) shows scenes of data assigned to the category (c2). The data of green objects are mostly included in this category. This indicates that “ao” was learned as a word to represent the color category related to green.
Similarly, Figure~\ref{fig:o7} shows Category (o7) in object modality associated with “ui” and scenes of data assigned to the category. From the graph of (a), it is confirmed that Category (o7) is associated with “ui.” Also, as shown in (b), the data of boxes are mostly included in this category. This indicates that “ui” was learned to represent the object category related to the box.
This shows that our proposed model can generate word sequences that represent novel situations by combinations of words associated with categories in specific modalities.

\end{CJK}

\section{Conclusions}\label{sec:7}
\begin{CJK}{UTF8}{min}
We presented a computational model for a symbol emergence system that allows the emergence of lexical knowledge with combinatoriality through category formation within individual agents and the exchange of word sequences among agents. Our proposed model, named Inter-CLS-PGM, was developed by extending the PGM of the M-H naming game~\cite{Inter-MDM-H2H} with the PGM of CSL~\cite{CSL}. 

To investigate whether our proposed model enables the emergence of lexical knowledge with combinatoriality among agents, we conducted two experiments in a simulation environment using humanoid robots. 
In the experiment of interpersonal cross-situational learning, two agents repeatedly formed categories and exchanged word sequences to learn the relationship between categories in multiple modalities and words in word sequences. We qualitatively evaluated whether our proposed model enables an agent to acquire lexical knowledge with combinatoriality and two agents to share words associated with categories in specific modalities and word sequences representing situations, compared to baseline models.
In the experiment of interpersonal cross-modal inference, an agent predicted the sensory-motor information of novel situations from word sequences proposed by another agent. We evaluated whether lexical knowledge facilitated by our proposed model demonstrated generalization performance to novel situations, compared to baseline models.

In the experimental results, we clarified the following.
\begin{itemize}
\item The mutual exclusivity constraint, which assigns novel words to unknown situational elements, plays a crucial role in the M-H naming game with CSL for the emergence of lexical knowledge with combinatoriality.
\item The M-H naming game enables the emergence of not only categorical signs but also lexical knowledge with combinatoriality dependent on modalities.
\item The lexical knowledge facilitated by our proposed model exhibits generalization performance to novel situations through interpersonal cross-modal inference.
\end{itemize}

As a future work, our proposed model, which handles word sequences, can be extended to a model that allows agents to communicate in natural language sentences. The ultimate goal of the series of research on symbol emergence systems~\cite{Taniguchi19, Taniguchi18}, including this study, is to elucidate how language emerged during human evolution and how it is acquired during human cognitive development.
By developing a model that employs natural language sentences for agent communication, a unique language will emerge among the agents. Moreover, new insights can be gained by analyzing this process. To achieve this goal, models for natural language processing (e.g., NPYLM~\cite{Mochihashi09} and combinatory categorial grammar~\cite{CCG}) will be integrated into our proposed model to extend it for learning word order and sentence structure.
We will conduct experiments of symbol emergence using daily objects with complex features, such as the YCB dataset~\cite{ycb} and the Fruits 360 dataset~\footnote{Fruits 360: https://www.kaggle.com/moltean/fruits}. Thus, we are developing a mutual learning model between neural network-based feature extraction using VAE~\cite{Kingma13,Srivastava17,razavi19}, Neuro-SERKET~\cite{Taniguchi20}, and PGM-based categorization and symbol emergence~\cite{Hagiwara18,Inter-MDM,Inter-MDM-H2H}.

\end{CJK}


\section*{Acknowledgments}
This study was partly supported by the Japan Society for the Promotion of Science (JSPS) KAKENHI under Grant JP21H04904 and JP18K18134, and MEXT Grant-in-Aid for Scientific Research on Innovative Areas 4903 (Co-creative Language Evolution), 17H06383.

\bibliographystyle{tfnlm}
\bibliography{main}

\clearpage

\appendix

\section{M-H naming game}
\label{apnd:naming game} 
\subsection{Models of M-H naming game}
Interpersonal Dirichlet mixtures (Inter-DM)~\cite{Inter-DM} is proposed as a PGM for category formation and symbol emergence among agents. In Inter-DM as a multi-agent extension of MLDA~\cite{Nakamura09}, modalities in MLDA are interpreted as agents and categories shared among modalities are interpreted as words shared among agents. The study argues that the inference algorithm based on Metropolis-Hastings method~\cite{Hastings70}, which generates a word from the proposal distribution and decides whether to accept or reject the word based on the acceptance ratio, can be interpreted as a naming game proposed by Steels et al~\cite{Steels15}. Experiments targeting daily objects using two robots with visual modality demonstrated that words associated with object categories are shared among agents in Inter-DM.

Interpersonal multimodal Dirichlet mixtures (Inter-MDM)~\cite{Inter-MDM} is a multimodal extension (i.e., vision, sound, and haptic) of Inter-DM with a single modality (i.e., vision). Experiments on symbol emergence among agents with different modalities demonstrated that the word exchange among agents improves the accuracy of categorization for each agent. Also, interpersonal cross-modal inference, in which Agent A proposes a word based on their own observation and Agent B predicts observations from the proposed word, was proposed in this study. However, because inter-MDM assumes a single word as a sign exchanged between agents, it cannot handle sentences composed of multiple words, as in human verbal communication.

Inter-MDM-H2H~\cite{Inter-MDM-H2H} is a model that has been extended to handle word sets composed of multiple words. This model can generate word sets as a Bag of Words representation from multimodal sensory information. However, Inter-MDM-H2H could not handle the combinatoriality of words because it infers one integrated category from multimodal sensory information.
For example, it could not generate the word sequence of "green cylinder" and its meaning by the combination of the words "green" associated with the color category and "cylinder" associated with the object category. 

\subsection{Generative process}

\begin{figure}[bt]
  \begin{center}			
  \includegraphics[scale=0.73]{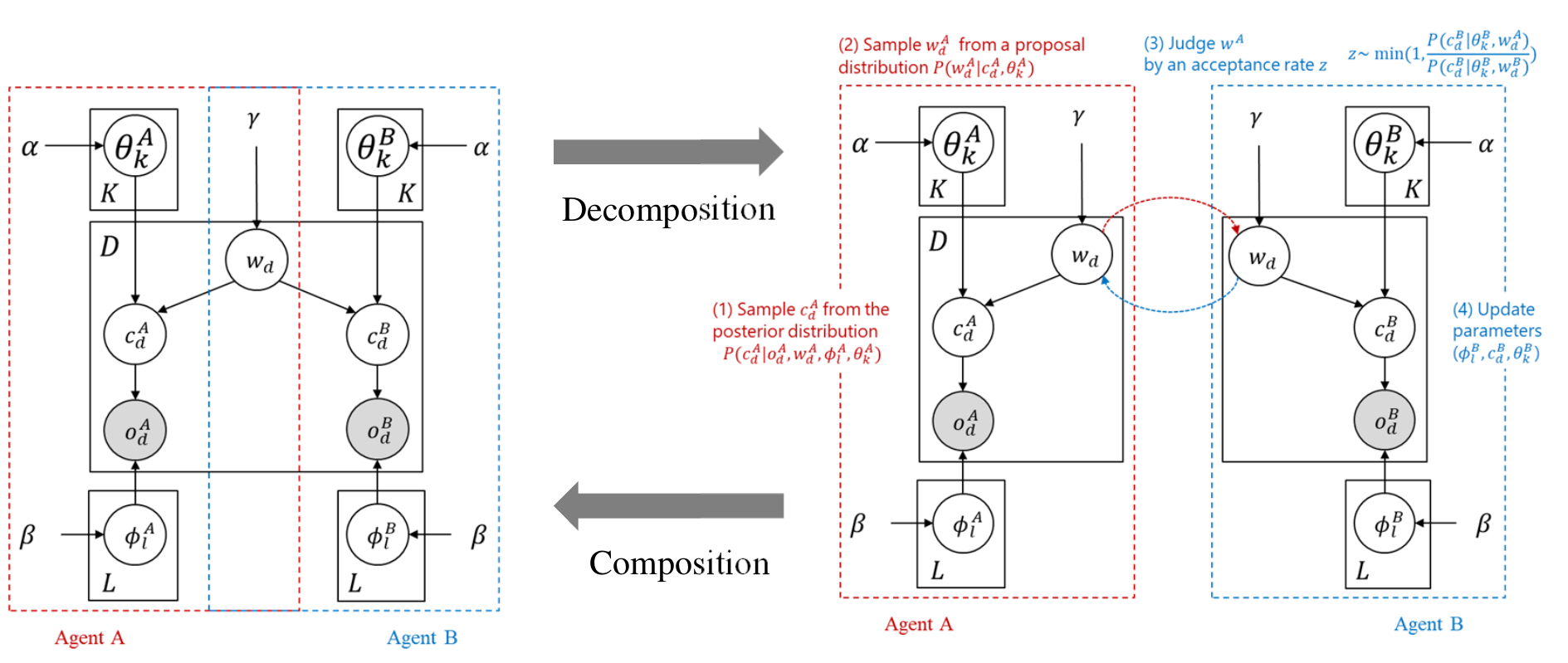}
  \caption{Probabilistic graphical model of M-H naming game and its decomposition.}
  \label{fig:serket}
  \end{center}
\end{figure}
Figure~\ref{fig:serket} illustrates a PGM of the M-H naming game as a simplified version of Inter-DM~\cite{Inter-DM} and its decomposition. In the graphical model, the left and right parts enclosed by dashed lines represent the models of agents A and B, respectively.
$w_{d}$ is a variable representing the word assigned to the data ($d$). The variable ($w_{d}$) can be interpreted as the word shared between agents A and B. $o_{d}^A$ and $o_{d}^B$ are the observed sensory-motor information of the data ($d$) in agents A and B. $c_{d}^A$ and $c_{d}^B$ are indices of the categories assigned to the data ($d$). $\phi_{l}^A$ and $\phi_{l}^B$ are the parameters of the probabilistic distribution for generating $o_{d}^A$ and $o_{d}^B$ with assigned categories ($c_{d}^A$, $c_{d}^B$). $\theta_{l}^A$ and $\theta_{l}^B$ are the parameters of the categorical distribution for  generating a category $c_{d}$ with an assigned word ($w_{d}$). $D$ is the number of data, $L$ is the number of categories, and $K$ is the number of words. $\alpha$,$\beta$,$\gamma$ are hyperparameters. 

\subsection{Inference as naming game}
The inference algorithm of the PGM is described as a naming game~\cite{Steels15} played by two agents, a speaker and a listener, as follows:
\begin{enumerate}
 \item The speaker estimates a category based on the observations obtained through their sensor.
 \item The speaker assigns a word based on an estimated category and sends it to the listener.
 \item The listener interprets the word sent by the speaker.
 \item The listener updates the knowledge of words associated with categories.
 \item Switch the speaker and the listener and repeat Processes (1)-(4).
\end{enumerate}

The inference algorithm was inspired by the symbol emergence in robotics tool KIT (SERKET)~\cite{SERKET, Taniguchi20}, a distributed development framework that enables the construction and inference of a large-scale PGM by connecting small-scale PGMs as sub-modules. By following SERKET, the independence of each module is preserved and the parameters of the entire model can be estimated with a minimum of parameter communication between modules. Experiments in the study~\cite{SERKET} demonstrated that the optimized parameters in a module-divided model following this framework are equivalent to those in an integrated model.

Figure~\ref{fig:serket} shows the model divided into two modules by SERKET. The integrated model on the left side is divided into two modules corresponding to agents A and B, which perform object categorization. The equations define the approximate inference of the connection variable $w_d$ between the modules based on the M-H algorithm.
Several mechanisms have been proposed in SERKET to communicate parameters between modules. The model employs a method based on the M-H algorithm. This method approximately infers $w_d$, a variable of the connection part, by the following procedure.

\begin{enumerate}
 \item Sample the index of a category ($c_d^A$) from the posterior distribution $P(c_d^A|o_{d}^A,{\bf{\Phi}}^A)$ based on the observed sensory-motor information ($o_{d}^A$) of Agent A.
 \item Sample a sign ($w_d^{A}$) from the proposed distribution $P(w_d^A|c_d^A,{\bf{\Theta}}^A)$ with the parameters of Agent A.
 \item Stochastically accept the sign ($w_d^{A}$) based on an acceptance ratio ($z$) calculated by the M-H algorithm in Agent B.
 \item Update the parameters of Agent B if the sign ($w_d^{A}$) is accepted.
 \item Switch agents and repeat Processes (1)-(4).
\end{enumerate}

This inference process can be interpreted as the naming game process described above and is called the M-H naming game.
Experiments in the study~\cite{Inter-DM} demonstrated that the repetition of the inference process leads to category formation based on observations and the emergence of signs associated with categories among agents.

\subsection{interpersonal cross-modal inference}
\label{sec:interpersonal cross-modal inference} 

\begin{figure}
  \begin{center}			
  \includegraphics[scale=0.6]{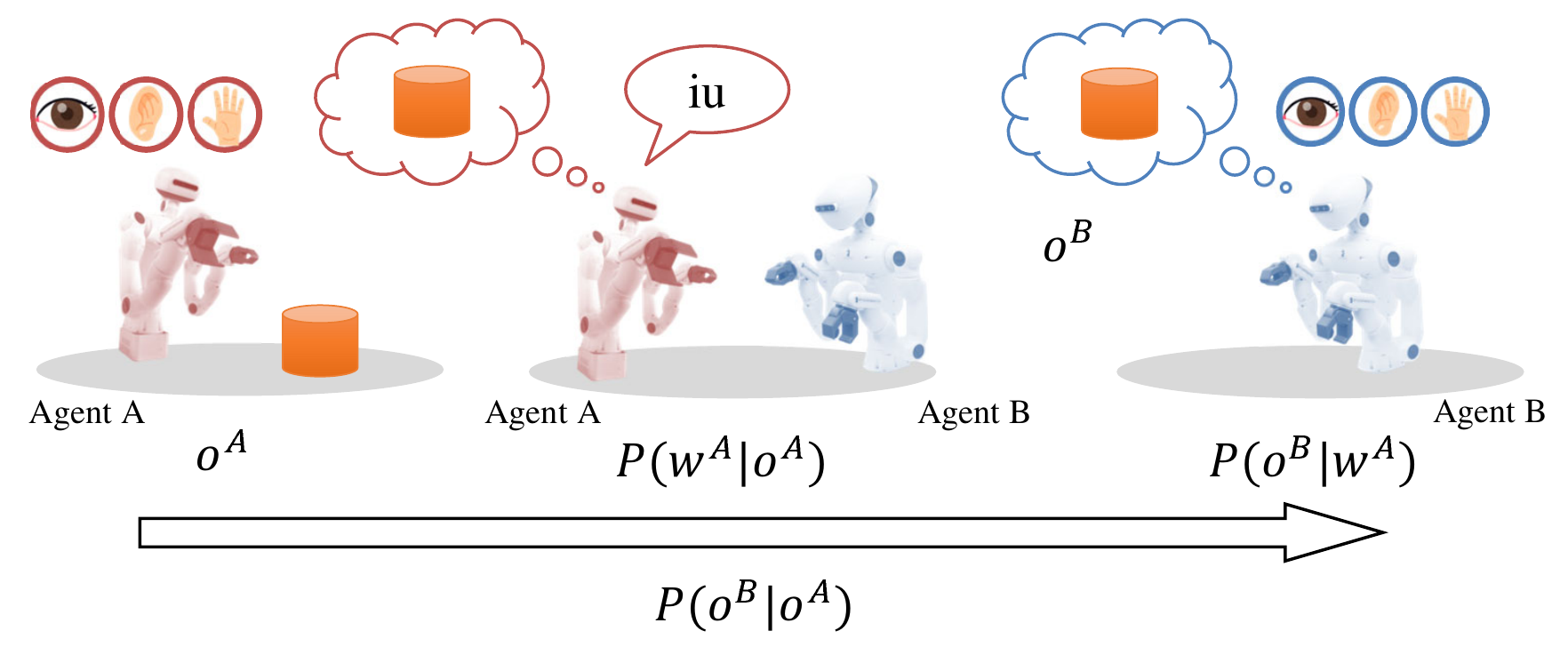}
  \caption{Overview of interpersonal cross-modal inference}
  \label{fig:predict_observation}
  \end{center}
\end{figure}

The interpersonal cross-modal inference is an important function of models for symbol emergence systems~\cite{Inter-DM,Inter-MDM,Inter-MDM-H2H}. In PGMs for object categorization based on multimodal information~\cite{Nakamura09,Muhammad13,Miyazawa19}, predicting the observation of one modality from the observation of another modality via categories is called cross-modal inference. For example, it enables an agent to predict unobserved tactile information from observed visual information about an apple via object categories. 
The interpersonal cross-modal inference is an extension of cross-modal inference from inter-modality to inter-agent. 
interpersonal cross-modal inference refers to inferring observations in one agent from observations in the other agent via words.
Figure~\ref{fig:predict_observation} shows an overview of interpersonal cross-modal inference.
For example, Agent A generates words corresponding to a category inferred from multimodal observations ($o^A$) of an object (e.g., orange cylinder) by the probability of $P(w^A|o^A)$. A generated word (e.g., ``iu'') is proposed to Agent B. Agent B predicts the multimodal observations based on the received words with the probability of $P(o^B|w^A)$. This process calculates the probability of $P(o^B|o^A)$ and implies the prediction of Agent B's observation ($o^B$) from Agent A's observation ($o^A$) via words ($w^A$). 
The interpersonal cross-modal inference is a computational model of the ability of one person to share words with another person and to understand the meaning of the words spoken by the other person by predicting modality information that they does not perceive. An example is the ability to imagine the color and shape of an apple when hearing the word "apple" spoken by another person. 
The interpersonal cross-modal inference is performed using fixed parameters in the model. 

\section{Derivation of the acceptance ratio in Inter-CSL-PGM} \label{apnd:acceptance ratio}
The derivation of the acceptance ratio ($r$) in Inter-CSL-PGM is described in the following equations. 
{\scriptsize
\begin{eqnarray}
\label{eq:acceptance ratio}
r&=&\frac{P\left(w_{dn}^{Sp}|F_d^{Sp},{\rm \bf{\Theta}}^{Sp},z_{**,d}^{Sp},F_d^{Li},{\rm \bf{\Theta}}^{Li},z_{**,d}^{Li}\right)P\left(w_{dn}^{Li}|F_d^{Sp},{\rm \bf{\Theta}}^{Sp},z_{**,d}^{Sp}\right)}
{P\left(w_{dn}^{Li}|F_d^{Sp},{\rm \bf{\Theta}}^{Sp},z_{**,d}^{Sp},F_d^{Li},{\rm \bf{\Theta}}^{Li},z_{**,d}^{Li}\right)P\left(w_{dn}^{Sp}|F_d^{Sp},{\rm \bf{\Theta}}^{Sp},z_{**,d}^{Sp}\right)} \nonumber \\
&\approx\propto&
\frac{P\left(w_{dn}^{Sp}|F_d^{Sp},{\rm \bf{\Theta}}^{Sp},z_{**,d}^{Sp}\right)
P\left(w_{dn}^{Sp}|F_d^{Li},{\rm \bf{\Theta}}^{Li},z_{**,d}^{Li}\right)
P\left(w_{dn}^{Li}|F_d^{Sp},{\rm \bf{\Theta}}^{Sp},z_{**,d}^{Sp}\right)}
{P\left(w_{dn}^{Li}|F_d^{Sp},{\rm \bf{\Theta}}^{Sp},z_{**,d}^{Sp}\right)
P\left(w_{dn}^{Li}|F_d^{Li},{\rm \bf{\Theta}}^{Li},z_{**,d}^{Li}\right)
P\left(w_{dn}^{Sp}|F_d^{Sp},{\rm \bf{\Theta}}^{Sp},z_{**,d}^{Sp}\right)}
\nonumber \\
&=&\frac{P\left(w_{nd}^{Sp}|F_d^{Li},{\rm \bf{\Theta}}^{Li},z_{**,d}^{Li}\right)}
{P\left(w_{dn}^{Li}|F_d^{Li},{\rm \bf{\Theta}}^{Li},z_{**,d}^{Li}\right)}
\nonumber \\
&=&\frac{P\left(w_{dn}^{Sp}|F_d^{Li},{\rm \bf{\Theta}}^{Li},z_{a,d}^{Li},z_{o,dA_d^{Li}}^{Li},z_{p,dA_d^{Li}}^{Li},z_{c,dA_d^{Li}}^{Li}\right)}
{P\left(w_{dn}^{Li}|F_d^{Li},{\rm \bf{\Theta}}^{Li},z_{a,d}^{Li},z_{o,dA_d^{Li}}^{Li},z_{p,dA_d^{Li}}^{Li},z_{c,dA_d^{Li}}^{Li}\right)},
\end{eqnarray}
}
where $z_{**,d}^{Sp}$ denotes $z_{a,d}^{Sp}$, $z_{o,dA_d^{Sp}}^{Sp}$, $z_{p,dA_d^{Sp}}^{Sp}$, $z_{c,dA_d^{Sp}}^{Sp}$, and $z_{**,d}^{Li}$ denotes $z_{a,d}^{Li}$, $z_{o,dA_d^{Li}}^{Li}$, $z_{p,dA_d^{Li}}^{Li}$, $z_{c,dA_d^{Li}}^{Li}$.
The product of experts~\cite{poe} is used for the transformation in Line 2 of Equations (\ref{eq:acceptance ratio}).

\section{Details of Inter-MDM-H2H-G} \label{apnd:details of inter-MDM-H2H-G}
\subsection{Overview}
\begin{figure}
  \begin{center}			
  \includegraphics[scale=0.7]{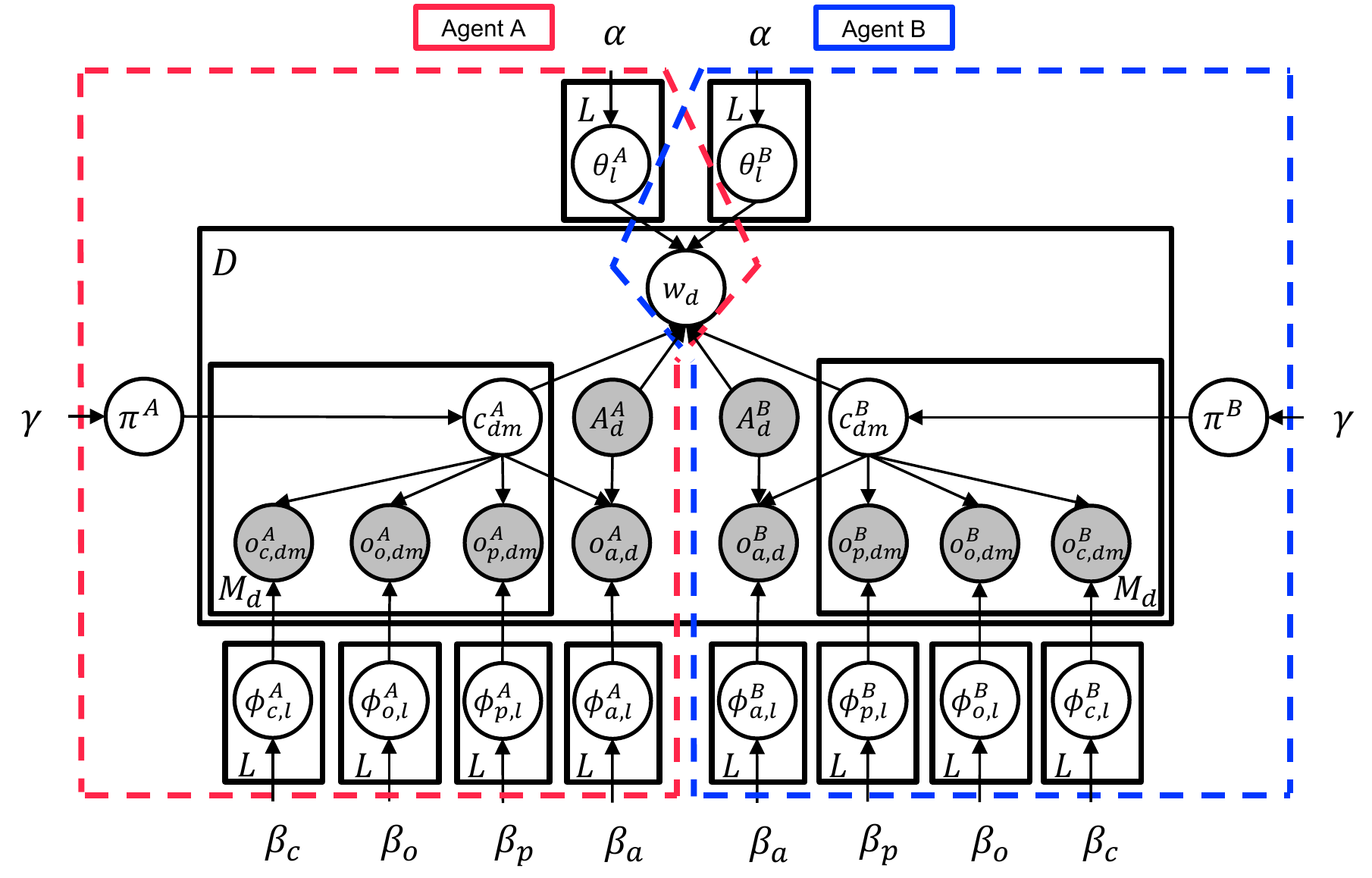}
  \caption{Graphical model of Inter-MDM-H2H-G}
  \label{fig:inter-MDM-H2H-G}
  \end{center}
\end{figure}

Figure~\ref{fig:inter-MDM-H2H-G} shows the graphical model Inter-MDM-H2H-G.
$w_{d}$ is the variable of words assigned to data ($d$).
$w_{d}$ can be interpreted as signs shared among Agents A and B.
$o_{*,dm}^A$ and $o_{*,dm}^B$ are the observed information of a object 
($m$) in data ($d$) in Agents A and B. 
$c_{dm}^A$ and $c_{dm}^B$ are the indices of the categories assigned to an object ($m$) in data ($d$). 
$\phi_{*,l}^A$ and $\phi_{*,l}^B$ are the parameters of the Gaussian distribution for generating $o_{*,dm}^A$ and $o_{*,dm}^B$ with assigned categories ($c_{dm}^A$, $c_{dm}^B$).
The $\pi^A$ and $\pi^B$ are the parameters of the categorical distribution for generating the categories ($c_{dm}^A$, $c_{dm}^B$).
$A_d^A, A_d^B$ is a variable for the index of the selected object ($m$) out of the objects ($M$) in each data in agents A and B.
$\theta_{l}^A$ and $\theta_{l}^B$ are the parameters of the multinomial distribution for generating signs $w_{d}$ with assigned categories ($c_{dA_d^A}^A$, $c_{dA_d^B}^B$).
$a$ is the action modality, $p$ is the position modality, $o$ is the object modality, and $c$ is the color modality. 
$D$ is the number of data, $M_d$ is the number of objects on the table
in the $d$-th data, and $L$ is the number of categories. $\alpha$,$\beta_*$,$\gamma$ are hyperparameters. 

In Inter-MDM-H2H-G, the probabilistic dependency between words ($w_{d}$) shared between Agents A and B and the categories ($c_{dm}^A$, $c_{dm}^B$) of each agent is a head-to-head type. The generative process of the head-to-head type allows for simple modeling of the exchange of word sets between agents using the Bag of Words representation.
As shown in the graphical model, observations ($o_{a,d}$,$o_{p,dm}$,$o_{o,dm}$,$c_{o,dm}$) of multiple modalities are generated from a single integrated category ($c_{dm}$) in each agent's model. 

\subsection{Generative process}
Since the structure of agents A and B is the same, the generation process of variables common among the agents is explained only for Agent A.
The generative process of Inter-MDM-H2H-G is described in Equations (\ref{eq:phi_A_h2h})-(\ref{eq:w_d_h2h}):
{\small
\begin{eqnarray}
\label{eq:phi_A_h2h}
\phi_{*,l}^A&\sim& {\rm GIW}(\beta_{*})\\
\label{eq:theta_A_h2h}
\theta_{l}^A&\sim& {\rm Dir}(\alpha)\\
\label{eq:pi_A_h2h}
\pi^A&\sim& {\rm Dir}(\gamma)\\
\label{eq:c_A_h2h}
c_{dm}^A&\sim& {\rm Cat}(\pi^A)\\
\label{eq:oa_A_h2h}
o_{a,d}^A&\sim& {\rm Gauss}(\phi_{a,c_{dA_d^A}^A}^A)\\
\label{eq:op_A_h2h}
o_{p,dm}^A&\sim& {\rm Gauss}(\phi_{p,c_{dm}^A}^A)\\
\label{eq:oo_A_h2h}
o_{o,dm}^A&\sim& {\rm Gauss}(\phi_{o,c_{dm}^A}^A)\\
\label{eq:oc_A_h2h}
o_{c,dm}^A&\sim& {\rm Gauss}(\phi_{c,c_{dm}^A}^A)\\
\label{eq:w_d_h2h}
w_{d}&\sim& \frac {{\rm Cat}(\theta_{c_{dA_d^A}^A}^A){\rm Cat}(\theta_{c_{dA_d^B}^B}^B)}{\sum_{w_{d}} {\rm Cat}(\theta_{c_{dA_d^A}^A}^A){\rm Cat}(\theta_{c_{dA_d^B}^B}^B)},
\end{eqnarray}
}
where Dir(.) denotes Dirichlet distribution, Gauss(.) denotes Gaussian distribution, Cat(.) denotes categorical distribution, and GIW(.) denotes Gaussian-inverse-Wishart distribution.
In Equation (\ref{eq:phi_A_h2h}), the parameter $\phi_{*,l}^A$ represents the probabilities for generating the features of observations in each modality ($*\in \{a,p,o,c\}$) at the plate($l\in L$).
In Equation (\ref{eq:theta_A_h2h}), $\theta_{l}^A$ represents the probabilities for generating a sign in each category at the plate ($l\in L$).
In Equation (\ref{eq:pi_A_h2h}), $\pi^A$ denotes the probabilities for generating categories.
In Equation (\ref{eq:c_A_h2h}), categories $c_{dm}^A$ at the plate $(d\in D)$ are generated from $\pi^A$.
In Equations (\ref{eq:oa_A_h2h})-(\ref{eq:oc_A_h2h}), observations ($o_{a,d}^A, o_{p,dm}^A, o_{o,dm}^A$ and $ o_{c,dm}^A$) are generated from the parameters ($\phi_{a,l}^A, \phi_{p,l}^A, \phi_{o,l}^A$ and $\phi_{c,l}^A$) with the categories ($c_{dm}^A$, $c_{dm}^B$).
In Equation (\ref{eq:w_d_h2h}), words ($w_{d}$) are generated from the parameters ($\theta_{l}^A$, $\theta_{l}^B$) with the categories ($c_{dm}^A$, $c_{dm}^B$).

\subsection{Inference algorithm}
The inference algorithm of Inter-MDM-H2H-G is described in Algorithm~\ref{alg:alg_h2h_2} based on M-H algorithm.
$\rm{\bf{C}}$ denotes the set of categories.
$c_d$ denotes $\{c_{d1}, c_{d2}, \cdots, c_{dM_d}\}$.
$i$ denotes the iteration. $\rm{\bf{O}}_{*}^A$ and $\rm{\bf{O}}_{*}^B$ denote the set of observations in each modality ($*$) for Agents A and B, respectively. 
$\rm{\bf{W}}$ denotes the set of words. 
In the algorithm, Agents A and B alternately sample the set of words $\rm{\bf{W}}$ and accept it based on the acceptance ratio based on the M-H algorithm.
In Line 3 of Algorithm~\ref{alg:alg_h2h_2}, the sampling and judgment of the set of words ${\rm \bf{W}}$ are performed with Agents A and B as the speaker and the listener, respectively.
In Line 4 of Algorithm~\ref{alg:alg_h2h_2}, sampling and judgment of the set of words ${\rm \bf{W}}$ are performed with Agent B as the speaker and Agent A as the listener.

Algorithm~\ref{alg:alg_h2h_3} denotes the M-H(.) part of Algorithm~\ref{alg:alg_h2h_2}. $Sp$ denotes the speaker, $Li$ denotes the listener, and Unif (.) denotes a uniform distribution.
The judgment refers to deciding to accept or reject a sampled word based on the acceptance rate $z$ (Line 4 of Algorithm~\ref{alg:alg_h2h_3}).

\begin{algorithm}
\caption{Inference algorithm for Inter-MDM-H2H-G}
\label{alg:alg_h2h_2}
    \begin{algorithmic}[1]
    \small
      \State Initialize all parameters
      \For{$i = 1$ to $I$}
        \State ${\bf W}^{B[i]},{\bf{C}}^{B[i]},{\bf{\Theta}}^{B[i]}$ = 
        \\M-H algorithm(${\bf{C}}^{A[i-1]},{\bf{\Theta}}^{A[i-1]},
        A_{d}^{A},
        {\bf{O}}^{B}_{a},{\bf{O}}^{B}_{p},{\bf{O}}^{B}_{o},{\bf{O}}^{B}_{c},A_{d}^{B},{\bf W}^{B[i-1]},{\bf{C}}^{B[i-1]},{\bf{\Theta}}^{B[i-1]}$)
        \State ${\bf W}^{A[i]},{\bf{C}}^{A[i]},{\bf{\Theta}}^{A[i]}$ = 
        \\M-H algorithm(${\bf{C}}^{B[i]},{\bf{\Theta}}^{B[i]},
        A_{d}^{B},
        {\bf{O}}^{A}_{a},{\bf{O}}^{A}_{p},{\bf{O}}^{A}_{o},{\bf{O}}^{A}_{c},A_{d}^{A},{\bf W}^{A[i-1]},{\bf{C}}^{A[i-1]},{\bf{\Theta}}^{A[i-1]}$)
      \EndFor
    \end{algorithmic}
\end{algorithm}

\begin{algorithm}
\caption{M-H algorithm in Inter-MDM-H2H-G}
\label{alg:alg_h2h_3}
\begin{algorithmic}[1]
\footnotesize
\State M-H algorithm$({\rm \bf{C}}^{Sp},{\rm \bf{\Theta}}^{Sp},A_{d}^{Sp},{\rm \bf{O}}_a^{Li},{\rm \bf{O}}_p^{Li},{\rm \bf{O}}_o^{Li},{\rm \bf{O}}_c^{Li},A_{d}^{Li},{\rm \bf{W}}^{Li},{\rm \bf{C}}^{Li},{\bf{\Theta}}^{Li})$:
\For {$d=1$ to $D$}
\State $w_{d}^{Sp} \sim P(w_{d}^{Sp}|c_{dA_d^{Sp}}^{Sp},{\bf{\Theta}}^{Sp})$
\State $z \leftarrow {\rm min}\left(1,
\dfrac{
P(w_{d}^{Sp}|c_{d}^{Li},{\bf{\Theta}}^{Li}, A_d^{Li})
}{
P(w_{d}^{Li}|c_{d}^{Li},{\bf{\Theta}}^{Li}, A_d^{Li})         
}
\right)$
\State $u \sim {\rm Unif}(0,1)$
\If {$u\leq z$}
\State $w_d = w_d^{Sp}$
\Else
\State $w_d = w_d^{Li}$
\EndIf
\EndFor
\For {$l=1$ to $L$}
\State $\phi_{a,l}^{Li} \sim {\rm GIW}(\phi_{a,l}^{Li}|{\rm \bf{O}}_a^{Li},{\rm \bf{C}}^{Li},\beta_{a}) $
\State $\phi_{p,l}^{Li} \sim {\rm GIW}(\phi_{p,l}^{Li}|{\rm \bf{O}}_p^{Li},{\rm \bf{C}}^{Li},\beta_{p}) $
\State $\phi_{o,l}^{Li} \sim {\rm GIW}(\phi_{o,l}^{Li}|{\rm \bf{O}}_o^{Li},{\rm \bf{C}}^{Li},\beta_{o}) $
\State $\phi_{c,l}^{Li} \sim {\rm GIW}(\phi_{c,l}^{Li}|{\rm \bf{O}}_c^{Li},{\rm \bf{C}}^{Li},\beta_{c}) $
\State  $\theta_{l}^{Li} \sim {\rm Dir}(\theta_{l}^{Li}|{\rm \bf{W}},{\rm \bf{C}}^{Li},\alpha) $
\EndFor
\State  $\pi^{Li} \sim {\rm Dir}(\pi^{Li}|{\rm \bf{C}}^{Li},\gamma) $
\For {$d=1$ to $D$}
\For {$m=1$ to $M_d$}
\State  $c_{dm}^{Li} \sim $ 
$
{\rm Cat}(c_{dm}^{Li}|\pi^{Li})
{\rm Cat}(w_{d}|\theta_{l=c_{dA_d^{Li}}}^{Li})$ 
\Statex \hspace{6em}$\times{\rm Gauss}(o_{a,d}^{Li}|\phi_{a,c_{dA_d^{Li}}^{Li}}^{Li})
{\rm Gauss}(o_{p,dm}^{Li}|\phi_{p,c_{dm}^{Li}}^{Li}) 
{\rm Gauss}(o_{o,dm}^{Li}|\phi_{o,c_{dm}^{Li}}^{Li}) 
{\rm Gauss}(o_{c,dm}^{Li}|\phi_{c,c_{dm}^{Li}}^{Li}) $
\EndFor
\EndFor
\State return {${\rm \bf{W}},{\rm \bf{C}}^{Li},{\bf{\Theta}}^{Li}$}
\end{algorithmic} 
\end{algorithm}

The acceptance rate ($z$) in the sampling from Agent A to B is calculated by Equation (\ref{eq:detail_z}) according to the equation of the acceptance rate in M-H algorithm.
The product of experts~\cite{poe} (i.e., $P(z|x,y) \approx\propto P(z|x)P(z|y)$) is used in the transformation of the equations from the Lines 1 to 2.

{\small
\begin{eqnarray}
\label{eq:detail_z}
z&=&\frac{P(w_d^{Sp}|c_d^{Sp},c_d^{Li},{\rm \bf{\Theta}}^{Sp},{\rm \bf{\Theta}}^{Li},A_d^{Li},A_d^{Sp})P(w_d^{Li}|c_d^{Sp},{\rm \bf{\Theta}}^{Sp},A_d^{Sp})}
{P(w_d^{Li}|c_d^{Sp},c_d^{Li},{\rm \bf{\Theta}}^{Sp},{\rm \bf{\Theta}}^{Li},A_d^{Li}, A_d^{Sp})P(w_d^{Sp}|c_d^{Sp},{\rm \bf{\Theta}}^{Sp},A_d^{Sp})} \nonumber \\
&\approx\propto&
\frac{P(w_d^{Sp}|c_d^{Sp},{\rm \bf{\Theta}}^{Sp},A_d^{Sp})P(w_d^{Sp}|c_d^{Li},{\rm \bf{\Theta}}^{Li},A_d^{Li})P(w_d^{Li}|c_d^{Sp},{\rm \bf{\Theta}}^{Sp},A_d^{Sp})}
{P(w_d^{Li}|c_d^{Sp},{\rm \bf{\Theta}}^{Sp},A_d^{Sp})P(w_d^{Li}|c_d^{Li},{\rm \bf{\Theta}}^{Li},A_d^{Li})P(w_d^{Sp}|c_d^{Sp},{\rm \bf{\Theta}}^{Sp},A_d^{Sp})}
\nonumber \\
&=&\frac{P(w_d^{Sp}|c_d^{Li},{\rm \bf{\Theta}}^{Li},A_d^{Li})}{P(w_d^{Li}|c_d^{Li},{\rm \bf{\Theta}}^{Li},A_d^{Li})} .
\end{eqnarray}
}

\subsection{Calculation process of interpersonal cross-modal inference} 
The calculation process of interpersonal cross-modal inference is described in Equations (\ref{eq:Cross_1})-(\ref{eq:Cross_7}).

{\small
\begin{eqnarray}
\label{eq:Cross_1}
c_{dA_d^{A}}^{A} &\sim&
P(c_{d}^{A}|o_{a,d}^{A},o_{p,dA_d^{A}}^{A},o_{o,dA_d^{A}}^{A},o_{c,dA_d^{A}}^{A},
{\bf{\Phi}}_{a}^{A},{\bf{\Phi}}_{p}^{A},{\bf{\Phi}}_{o}^{A},{\bf{\Phi}}_{c}^{A}) \\
\label{eq:Cross_2}
w_{d}^{A} &\sim& P(w_{d}^{A}|c_{dA_d^{A}}^{A},{\bf{\Theta}}^{A}) \\
\label{eq:Cross_3}
c_{dA_d^{B}}^{B} &\sim&
P(c_{dA_d^{B}}^{B}|w_d^A,{\bf{\Theta}}^{B}) \\
\label{eq:Cross_4}
\hat{o_{a,d}^{B}} &\sim&
P(\hat{o_{a,d}^{B}}|c_{dA_d^{B}}^{B},{\bf{\Phi}}_{a}^{B}) \\
\label{eq:Cross_5}
\hat{o_{p,dA_d^{B}}^{B}} &\sim&
P(\hat{o_{p,dA_d^{B}}^{B}}|c_{dA_d^{B}}^{B},{\bf{\Phi}}_{p}^{B}) \\
\label{eq:Cross_6}
\hat{o_{o,dA_d^{B}}^{B}} &\sim&
P(\hat{o_{o,dA_d^{B}}^{B}}|c_{dA_d^{B}}^{B},{\bf{\Phi}}_{o}^{B}) \\
\label{eq:Cross_7}
\hat{o_{c,dA_d^{B}}^{B}} &\sim&
P(\hat{o_{c,dA_d^{B}}^{B}}|c_{dA_d^{B}}^{B},{\bf{\Phi}}_{c}^{B}),
\end{eqnarray}
}

where $\hat{o_{a,d}^{B}}$, $\hat{o_{p,dA_d^B}^{B}}$, $\hat{o_{o,dA_d^B}^{B}}$, and $\hat{o_{c,dA_d^B}^{B}}$ represent predicted observations in modalities (i.e., action, position, object, color) of agent B, respectively.
${\bf{\Phi}}_{*}^{A}, {\bf{\Phi}}_{*}^{B}$ denote set of parameters $(\phi_{*,l}^{A}, \phi_{*,l}^{B})$.

Equation (\ref{eq:Cross_1}) is interpreted as the process where Agent A infers Category $c_{dA_d^{A}}^A$ from its own observations ($o_{*,dA_d^A}^A$) and parameters ($\phi_*^A$).
Equation (\ref{eq:Cross_2}) is interpreted as the process where Agent A proposes words $w_d^A$ based on the inferred Category $c_{dA_d^{A}}^A$.
Equation (\ref{eq:Cross_3}) is interpreted as the process where Agent B infers Category $c_{dA_d^{B}}^B$ based on the proposed words $w_d^A$ and its own parameter ${\bf{\Theta}}^{B}$.
Equations (\ref{eq:Cross_4})-(\ref{eq:Cross_7}) are interpreted as the process where Agent B predicts observations in each modality from the inferred Category $c_{dA_d^{B}}^B$.


\end{document}